\documentclass[11pt]{article}

\usepackage[final]{acl}

\usepackage{times}
\usepackage{latexsym}

\usepackage[T1]{fontenc}
\usepackage[utf8]{inputenc}

\usepackage{microtype}

\usepackage{inconsolata}
\usepackage{amsmath}
\usepackage{amssymb}

\usepackage{graphicx}

\usepackage{booktabs}
\usepackage{multirow}
\usepackage{colortbl}
\usepackage{listings}
\usepackage{subcaption}
\usepackage{cleveref}
\usepackage{placeins}

\usepackage[table]{xcolor}
\usepackage{arydshln}
\newcolumntype{H}{>{\setbox0=\hbox\bgroup}c<{\egroup}}
\definecolor{bluerow}{HTML}{E9F2FB}

\title{How to Score Experts for One-Shot MoE Expert Pruning: A Systematic Formulation and Selection Principle}

\author{
  \textbf{Zongfang Liu\textsuperscript{1,2}}
  \quad
  \textbf{Jinghui Zhang\textsuperscript{4}}
  \quad
  \textbf{Zijian Ma\textsuperscript{1,2}}
  \quad
  \textbf{Guangyi Chen\textsuperscript{3,4,\dag}}
  \quad
  \textbf{Xin Yuan\textsuperscript{2,\dag}}
\\
\\
  \normalfont
  \textsuperscript{1}Zhejiang University, Hangzhou, China
  \quad
  \textsuperscript{2}Westlake University, Hangzhou, China
\\
  \textsuperscript{3}Carnegie Mellon University, Pittsburgh, PA, USA
\\
  \textsuperscript{4}Mohamed bin Zayed University of Artificial Intelligence, Abu Dhabi, UAE
\\
  \small{\textsuperscript{\dag}\textbf{Corresponding authors:}
  \href{mailto:guangyichen1994@gmail.com}{\texttt{guangyichen1994@gmail.com}},
  \href{mailto:xyuan@westlake.edu.cn}{\texttt{xyuan@westlake.edu.cn}}.}
}

\begin{document}
\maketitle
\begin{abstract}
Mixture-of-Experts (MoE) language models reduce per-token computation through sparse expert activation, yet deployment still requires storing the full expert pool, making one-shot expert pruning a practical approach for reducing memory usage. Although effective, existing criteria are largely heuristic, and no single criterion is universally optimal. Thus, establishing a principle for selecting pruning criteria suited to different deployment objectives remains an important yet largely underexplored problem in one-shot expert pruning.
To this end, we introduce a systematic formulation for one-shot MoE expert pruning organized around three factors: routing frequency, gate weighting, and activation strength. The formulation yields a criteria selection principle: task-agnostic pruning should favor routed-token-averaged, gate-free activation-based criteria, whereas task-specific pruning can benefit from retaining routing-frequency and gate-weight information. Beyond this principle, the formulation also provides a systematic view of existing heuristic criteria and gives rise to two new task-agnostic criteria, Mean Activation Norm (MAN) and Mean Squared Activation Norm (MSAN). Across four representative MoE models and 16 diverse benchmarks, MAN and MSAN are consistently strong in the task-agnostic setting, obtain the top-two average ranks, and improve average performance by 4.0 points on average and up to 8.8 points over the strongest baseline.
\end{abstract}

\section{Introduction}

Mixture-of-Experts (MoE) language models scale capacity through sparse conditional computation, replacing dense feed-forward blocks with expert pools from which a router selects only a few experts per token \cite{shazeer2017outrageously,fedus2022switch}. This mechanism has become a central scaling strategy for recent language models \cite{jiang2024mixtral,muennighoff2024olmoe,liu2024deepseek,yang2025qwen3,baidu2025ernie45,meta2025llama,glm5team2026glm5vibecodingagentic,kimiteam2026kimik25visualagentic,qwen3.5}. However, sparse MoE primarily reduces computation per token, while the memory footprint remains tied to the full expert pool: all experts must still be stored, loaded, and managed during inference. Consequently, memory usage becomes the major deployment bottleneck for MoE models. This motivates a growing line of expert-level compression methods~\cite{li2023merge,lu2024not,zhang2025diversifying,lee2025stun}. Among these, one-shot expert pruning is attractive because it ranks and prunes experts in a single calibration pass, without finetuning, retraining, or extensive combinatorial search~\cite{muzio2024seer,jaiswal2025finding,lasby2026reap}. 

\begin{figure}[t]{
  \centering
  \includegraphics[width=\columnwidth]{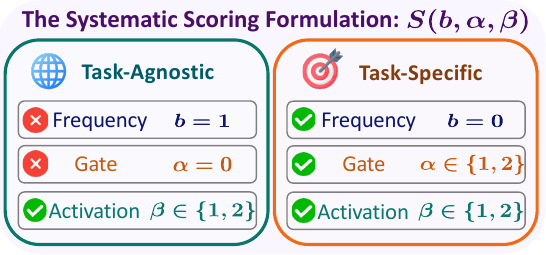}
  \caption{\textbf{Selection principle derived from the systematic formulation $\mathcal{S}(b,\alpha,\beta)$.} Task-agnostic pruning favors routed-token-averaged, gate-free activation criteria, whereas task-specific pruning can benefit from routing-frequency and gate-weight information. 
  }
  \label{fig:principle}
  }
  \vspace{-1em}
\end{figure}

\begin{figure}[t]
  \centering
  \includegraphics[width=\columnwidth]{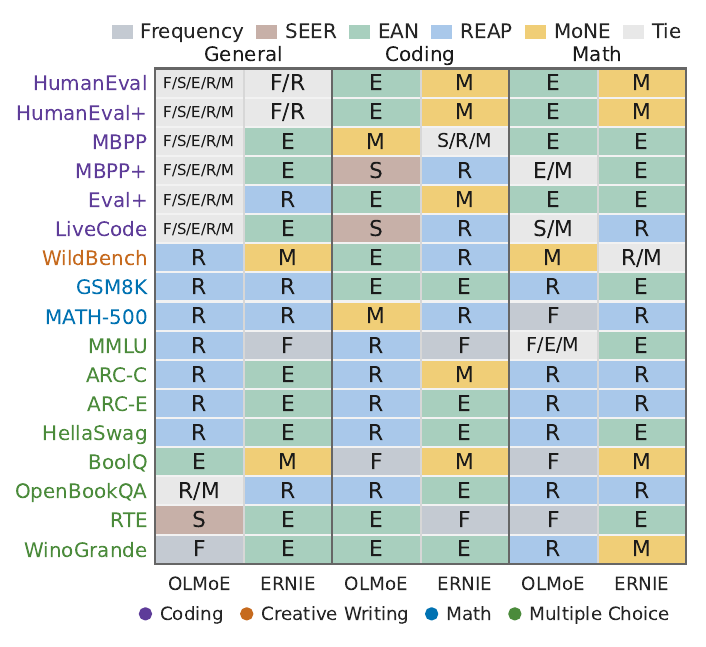}
  \caption{\textbf{Winner map for pruning scoring criteria under different calibration sets and models.} Each cell reports the scoring criterion with the highest benchmark score among Frequency, SEER, EAN, REAP, and MoNE for OLMoE-7B at a 25\% pruning ratio and ERNIE-4.5-21B at a 50\% pruning ratio. Columns group results by calibration set: C4 (General), Evol-CodeAlpaca-v1 (Coding), and Tulu-3-SFT-Personas-Math (Math). No single scoring criterion is universally optimal across these settings.}
  \label{fig:motivation_panel_a}
  \vspace{-1em}
\end{figure}

While one-shot expert pruning is simple and effective, existing methods largely rely on heuristic expert-importance criteria proposed in isolation, and as reflected in \Cref{fig:motivation_panel_a}, no single criterion is consistently optimal across calibration sets, evaluation tasks, and models.
This inconsistency likely arises from the fact that one-shot expert pruning is fundamentally a tradeoff. Without retraining or finetuning, preserving experts that support one capability may require discarding experts that support another, especially at high pruning ratio~\cite{liu2026aimercalibrationfreetaskagnosticmoe}. Moreover, different scoring criteria encode different notions of expert importance. Some are more closely tied to the calibration set, whereas others may capture more stable patterns of expert utility that better transfer across tasks~\cite{zhang2026monereplacingredundantexperts}. Thus, rather than assuming the existence of a single universally optimal pruning criterion, we identify criterion selection as a central yet underexplored problem in one-shot expert pruning. This raises a natural question: \textit{how should appropriate pruning criteria be selected for different deployment objectives?}

To this end, we first use a single-expert pruning damage measure to identify three core components underlying one-shot expert-pruning criteria: routing frequency, gate weighting, and activation strength. Based on these components, we provide a systematic formulation that characterizes how each component affects calibration dependence and derive a criteria selection principle, summarized in \Cref{fig:principle}. This formulation also places existing criteria such as Frequency, SEER~\cite{muzio2024seer}, EAN~\cite{jaiswal2025finding}, and REAP~\cite{lasby2026reap} into a common design space, explaining why different criteria are suitable for different pruning objectives. Guided by the proposed principle, we further derive two task-agnostic criteria, Mean Activation Norm (MAN) and Mean Squared Activation Norm (MSAN). Across four representative MoE models and 16 downstream benchmarks spanning coding, creative writing, mathematical reasoning, and multiple-choice question answering, the proposed criteria achieve the top-two average ranks and improve average performance by 4.0 points on average and up to 8.8 points over the strongest prior baseline. The accompanying implementation is publicly available.\footnote{\url{https://github.com/ZongfangLiu/unified-expert-pruning}}

\noindent\textbf{Our contributions are summarized as follows:}
\begin{itemize}
    \vspace{-0.4em}
    \item By analyzing single-expert pruning damage, we identify three core components for one-shot expert pruning criteria: routing frequency, gate weighting, and activation strength.
    \item Based on these components we provide a systematic scoring formulation for one-shot expert pruning and derive a clear principle for task-agnostic and task-specific pruning.
    \item Guided by this principle, we derive two task-agnostic criteria, MAN and MSAN. Experiments on four MoE models and 16 diverse benchmarks show that the proposed criteria deliver more balanced overall performance.
\end{itemize}

\section{Related Work}
\subsection{Expert Pruning}
Early work on MoE expert pruning considers downstream task specialization and show that substantial expert redundancy can be removed after task-specific fine-tuning~\cite{chen2022task}. \citet{koishekenov2023memory} study multilingual machine translation and show that pruning language-specific experts improves memory efficiency while largely preserving performance. NAEE~\cite{lu2024not} selects retained experts by minimizing the Frobenius-norm reconstruction error between original and pruned layer outputs. \citet{bai2025diepadaptivemixtureofexpertscompression, liu2024efficient,liu2026evoesap} formulate pruning as an optimization problem, using gradient-based or gradient-free procedures to optimize expert importance or layer-wise pruning-ratio allocation. EASY-EP~\cite{dong2025domainspecificpruninglargemixtureofexperts} identifies domain-relevant experts from a few in-domain demonstrations. STUN~\cite{lee2025stun} clusters experts by router behaviors to prune or merge redundant ones and then applies unstructured pruning to remaining experts. HodgeCover~\cite{zhong2026hodgecover} performs compression using higher-order topological coverage. Beyond these methods, one-shot expert pruning has emerged as a practical and actively studied approach, owing to its simplicity and effectiveness. Existing methods mainly differ in the expert-importance scoring criterion they adopt. SEER-MoE~\cite{muzio2024seer} ranks experts by routing frequency (Frequency) or gate-weighted routing frequency (SEER). \citet{jaiswal2025finding} compares different scoring criteria and identify the sum of expert activation norm (EAN) as the strongest. REAP~\cite{lasby2026reap} uses gate-weighted activation norms and achieves strong performance on generative tasks, while MoNE~\cite{zhang2026monereplacingredundantexperts} uses routing-frequency-weighted activation variance to capture expert redundancy. Despite their practical appeal, existing one-shot expert-pruning methods mainly seek stronger individual scoring criteria. Yet no single scoring criterion consistently dominates, since one-shot pruning inherently trades off different capabilities. Our work differs in taking a systematic view of these criteria, relating them through a common score family and characterizing which choices are more suitable for task-specific and task-agnostic pruning.

\subsection{Expert Merging}
Expert merging compresses MoE models by consolidating multiple experts into fewer experts. MEO~\cite{he2023merging} performs online token-wise merging by forming a router-score-weighted combination of the activated experts at inference time. Offline methods instead merge experts ahead of deployment: MC-SMoE~\cite{li2023merge} first aligns neurons, groups experts using routing information, and merges each group with routing-frequency-weighted averaging; HC-SMoE~\cite{chen2024retraining} hierarchically clusters experts by output similarity before merging each cluster; and Sub-MoE~\cite{li2025sub} clusters experts and merges them in a shared subspace. REAM~\cite{jha2026reammergingimprovespruning} is a REAP-inspired variant that groups experts and merges their weights instead of pruning them. 

\subsection{Other Compression Methods}
Beyond whole-expert pruning and merging, SMoE models can also be compressed at a finer granularity through quantization~\cite{huang2024mixture}, decomposition-based compression~\cite{gu2025delta,he2025efficiently,li2025moe}, and intra-expert weight pruning as in MoE-Pruner~\cite{xie2024moe}. Some methods also combine expert-level compression with finer-grained reconstruction. DERN~\cite{zhou2025dropping} first prunes redundant experts and then reallocates neuron-level segments to retained experts. Another line of work combines multiple techniques~\cite{he2024towards,liu2024survey}. For example, MoE-I$^2$~\cite{yang2024moe} integrates expert pruning and low-rank decomposition, followed by LoRA fine-tuning~\cite{hu2022lora} to recover performance.

\section{Preliminaries}
\noindent\textbf{Mixture-of-Experts Layer.}
A Mixture-of-Experts (MoE) layer consists of $n$ feed-forward networks, referred to as experts $\{E_i\}_{i=1}^{n}$, and a router that activates only the top-$k$ experts for each token~\cite{shazeer2017outrageously,fedus2022switch}. For a token $t$ with hidden representation $\mathbf{h}_t\in\mathbb{R}^{d}$, the router computes logits $\mathbf{z}_t=\mathbf{W}_r \mathbf{h}_t \in \mathbb{R}^{n}$,
where $\mathbf{W}_r \in \mathbb{R}^{n\times d}$ is the router projection matrix. Let
$\mathcal{E}_t:=\mathrm{TopK}(\mathbf{z}_t,k)$ denote the set of selected experts, where $k\ll n$. The router then normalizes the logits over $\mathcal{E}_t$, yielding sparse gate weights
\begin{equation}
g_{i,t}=
\begin{cases}
\dfrac{\exp(z_{i,t})}{\sum_{j\in \mathcal{E}_t}\exp(z_{j,t})}, & i\in \mathcal{E}_t,\\[6pt]
0, & i\notin \mathcal{E}_t.
\end{cases}
\label{eq:topk_gate}
\end{equation}
Let $\mathbf{f}_{i,t}:=E_i(\mathbf{h}_t)\in\mathbb{R}^{d}$ denote the output of expert $i$ for token $t$. The MoE layer output is
\begin{equation}
\mathbf{y}_t=\textstyle \sum_{i\in \mathcal{E}_t} g_{i,t}\,\mathbf{f}_{i,t}.
\label{eq:moe_output}
\end{equation}
\noindent\textbf{One-shot expert pruning.}
Given a calibration set, one-shot expert pruning assigns an importance score to the experts in each layer, ranks them accordingly, and prunes the least important ones to meet a target pruning ratio without any finetuning, retraining, or extensive combinatorial search.

\section{Methodology}
\subsection{Damage Measure for Expert Removal}
For an MoE layer, consider the pruned configuration in which the \(j\)-th expert
\(E_j\) and its associated router parameters are removed. The router then computes
logits over the remaining experts, followed by top-\(k\) selection and gate
normalization. Let \(\mathcal{E}_t^{(-j)}\) and
\(\tilde{g}_{i,t}^{(-j)}\) denote the selected expert set and gate weights after
this removal, respectively. For a token originally routed to \(E_j\),
removing \(E_j\) promotes a replacement expert \(E_r\). The original layer output can
be decomposed by separating the contribution of \(E_j\) from those of the
other active experts:
\begin{equation}
\mathbf{y}_t
=
\textstyle g_{j,t}\mathbf{f}_{j,t}
+
\sum_{i\in \mathcal{E}_t\setminus\{j\}}g_{i,t}\mathbf{f}_{i,t}.
\label{eq:yt_split_j}
\end{equation}
After \(E_j\) is removed, the experts that remain active are reweighted,
and the replacement expert \(E_r\) is added:
\begin{equation}
\mathbf{y}_t^{(-j)}
=
\textstyle \sum_{i\in \mathcal{E}_t\setminus\{j\}}
\tilde{g}_{i,t}^{(-j)}\mathbf{f}_{i,t}
+
\tilde{g}_{r,t}^{(-j)}\mathbf{f}_{r,t}.
\label{eq:yt_pruned_split_j}
\end{equation}
Subtracting \Cref{eq:yt_pruned_split_j} from \Cref{eq:yt_split_j} gives the exact output
perturbation induced by pruning \(E_j\):
\begin{equation}
\Delta_{j,t}
=
\mathbf{y}_t-\mathbf{y}_t^{(-j)}
=
g_{j,t}\mathbf{f}_{j,t}
+
\boldsymbol{\rho}_{j,t},
\label{eq:delta_component_summary}
\end{equation}
where \(g_{j,t}\mathbf{f}_{j,t}\) is the direct contribution removed with \(E_j\),
and \(\boldsymbol{\rho}_{j,t}\) collects the rerouting effects:
\begin{equation}
\boldsymbol{\rho}_{j,t}
:=
\textstyle \sum_{i\in \mathcal{E}_t\setminus\{j\}}
\left(g_{i,t}-\tilde{g}_{i,t}^{(-j)}\right)\mathbf{f}_{i,t}
-
\tilde{g}_{r,t}^{(-j)}\mathbf{f}_{r,t},
\label{eq:rerouting_residual}
\end{equation}
where the first term captures gate renormalization on the
surviving experts, and the second term captures the contribution of the promoted
replacement expert. Thus, \Cref{eq:delta_component_summary} separates the output change
into a removed-expert contribution and a rerouting residual. 

A scalar damage score can be obtained by measuring the norm of the output
perturbation. Given a calibration set with \(M\) tokens, the exact single-expert damage can then be written as
\begin{equation}
\widehat{D}^{\Delta}_j
:=
\textstyle \sum_{t=1}^{M}
\mathbf{1}[j\in \mathcal{E}_t]\,
\|\Delta_{j,t}\|_2.
\label{eq:exact_damage_score}
\end{equation}
Directly evaluating \(\widehat{D}^{\Delta}_j\) requires simulating the removal of each
candidate expert \(E_j\): the router logits for \(E_j\) must be masked, the top-\(k\) set
\(\mathcal{E}_t^{(-j)}\) must be recomputed, and the selected gates must be renormalized to
obtain \(\tilde{g}_{i,t}^{(-j)}\). Repeating this procedure for every expert and token is more expensive than a standard calibration pass. In one-shot expert pruning,
the importance scores are expected to be collected once on a calibration set and then used
directly for ranking, without an expert-wise rerouting procedure. A direct proxy is
therefore to ignore the rerouting residual \(\boldsymbol{\rho}_{j,t}\) in
\Cref{eq:delta_component_summary} and measure only the removed contribution
\(g_{j,t}\mathbf{f}_{j,t}\), giving
\begin{equation}
\textstyle \widehat{D}_j
:=
\sum_{t=1}^{M}
\mathbf{1}[j\in \mathcal{E}_t]\,
g_{j,t}\|\mathbf{f}_{j,t}\|_2 .
\label{eq:damage_proxy}
\end{equation}

By \Cref{eq:delta_component_summary} and the triangle inequality, the
approximation error satisfies
\begin{equation}
\left|\widehat{D}^{\Delta}_j-\widehat{D}_j\right|
\leq
\sum_{t=1}^{M}
\mathbf{1}[j\in\mathcal{E}_t]\,
\|\boldsymbol{\rho}_{j,t}\|_2.
\label{eq:proxy_error_bound}
\end{equation}

\begin{figure}[t]
  \centering
  \includegraphics[width=\columnwidth]{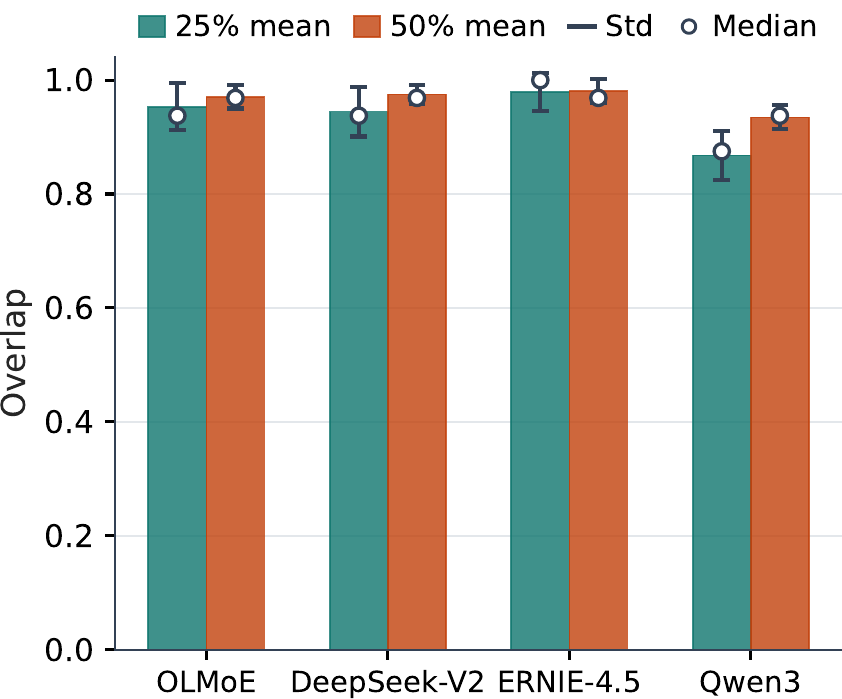}
  \vspace{-6mm}
  \caption{\textbf{Expert overlap between the single-expert proxy and exact damage under 25\% and 50\% pruning ratios.} We use \(\sim\)0.5M tokens from C4, and then rank the experts within each layer by the proxy damage score \(\widehat{D}_j\) and the exact damage score \(\widehat{D}^{\Delta}_j\). Across four distinct and representative models, the overlap is mostly close to or larger than 0.95, and even the lowest case, Qwen3 at 25\%, remains approximately above 0.85.}
  \label{fig:single_expert_delta}
  \vspace{-2mm}
\end{figure}

Empirically, \Cref{fig:single_expert_delta} indicates that the expert ranking induced by the proxy damage score is closely aligned with that induced by the exact damage score. This makes \(\widehat{D}_j\) a practical choice for one-shot expert pruning, as it avoids the expert-wise recomputation needed to evaluate \(\widehat{D}^{\Delta}_j\).

\begin{figure*}[t]
  \centering
  \newsavebox{\figthreea}
  \newsavebox{\figthreeb}
  \newlength{\figthreelabelwidth}
  \setlength{\figthreelabelwidth}{1.8em}
  \begin{subfigure}{\textwidth}
    \centering
    \sbox{\figthreea}{\includegraphics[width=0.98\dimexpr\textwidth-\figthreelabelwidth\relax]{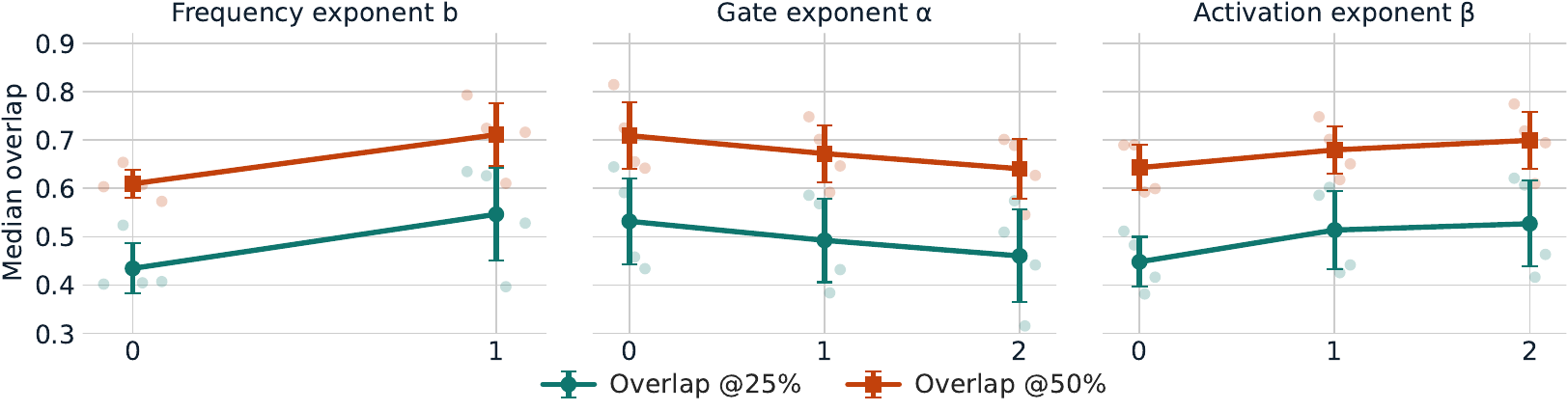}}
    \makebox[\figthreelabelwidth][l]{\hspace{0.3em}\raisebox{0.5\ht\figthreea}{(a)}}%
    \usebox{\figthreea}
    \phantomcaption\label{fig:unified_score_components_a}
  \end{subfigure}

  \vspace{0.8em}
  \begin{subfigure}{\textwidth}
    \centering
    \sbox{\figthreeb}{\includegraphics[width=0.96\dimexpr\textwidth-\figthreelabelwidth\relax]{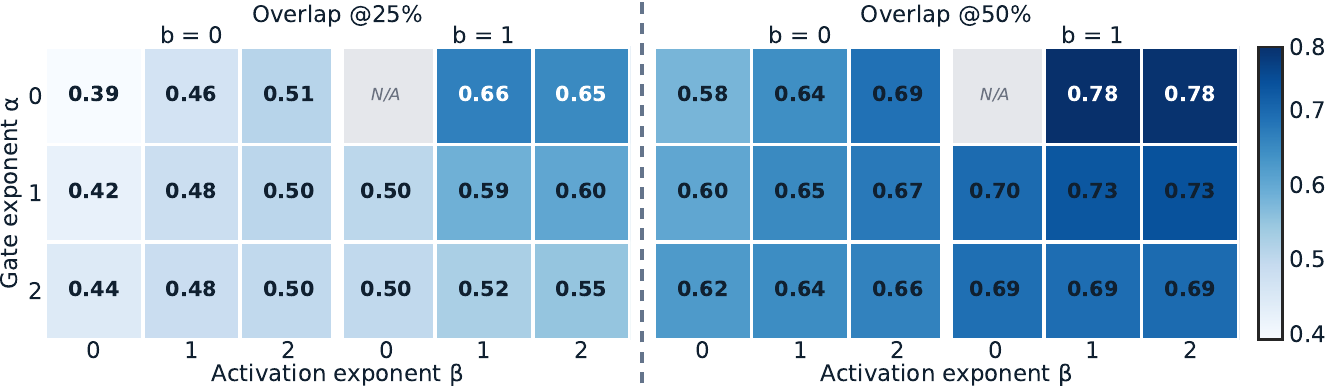}}
    \makebox[\figthreelabelwidth][l]{\hspace{0.3em}\raisebox{0.5\ht\figthreeb}{(b)}}%
    \usebox{\figthreeb}
    \phantomcaption\label{fig:unified_score_components_b}
  \end{subfigure}
  \caption{\textbf{Overlap of bottom-ranked experts across calibration distributions for different score variants.} Panel~(\subref{fig:unified_score_components_a}): Individual component effects. Panel~(\subref{fig:unified_score_components_b}): Score-variant trends.
For each score variant $S_j(b,\alpha,\beta)$, we compute expert rankings using \(\sim\) 0.5M training tokens from each of three calibration sets: C4, Evol-CodeAlpaca-v1, and Tulu-3-SFT-Personas-Math. For each model, we then measure the overlap among the bottom-ranked experts selected for pruning under the three calibration sets at 25\% and 50\% pruning ratios. N/A denotes the degenerate case $(1,0,0)$, which yields a constant score. Results are averaged over the four models: OLMoE-1B-7B-0125-Instruct, DeepSeek-V2-Lite-Chat, ERNIE-4.5-21B-A3B-PT, and Qwen3-30B-A3B-Instruct-2507.
}
  \label{fig:unified_score_components}
  \vspace{-1em}
\end{figure*}

\subsection{The Systematic Scoring Formulation}
\noindent\textbf{Routing frequency, gate weights and activation strength.}
The proxy damage score in \Cref{eq:damage_proxy} exposes three factors that recur across existing expert-importance scoring criteria. The indicator $\mathbf{1}[j\in\mathcal{E}_t]$ determines which tokens are routed to \(E_j\) and therefore carries routing-frequency information. The gate weight \(g_{j,t}\) measures how strongly the router assigns weight to \(E_j\) on those tokens. The activation term $\|\mathbf{f}_{j,t}\|_2$ measures the strength of the expert output. Because these factors appear multiplicatively in \(\widehat{D}_j\), we can ablate or emphasize each factor directly.

\noindent\textbf{The systematic formulation.}
Based on these factors, we define a systematic scoring formulation for one-shot expert pruning:
\begin{equation}
\textstyle S_j(b,\alpha,\beta)
:=
 \frac{1}{N_j^b}
\sum_{t=1}^{M}
\mathbf{1}[j\in \mathcal{E}_t]\,
g_{j,t}^{\alpha}
\|\mathbf{f}_{j,t}\|_2^{\beta},
\label{eq:score_family}
\end{equation}
where \(N_j := \sum_{t=1}^{M}\mathbf{1}[j\in\mathcal{E}_t]\) is the number of routed tokens for expert \(j\), and the hyperparameters satisfy \(b\in\{0,1\}\) and \(\alpha,\beta\in\{0,1,2\}\). Here \(b\) determines whether routing frequency is retained (\(b\)=\(0\)) or converted to a routed-token average (\(b\)=\(1\)); \(\alpha\) is the gate-weight exponent, with \(\alpha\)=\(0\) removing gate weighting and larger \(\alpha\) making the score more sensitive to gate magnitude by emphasizing routed tokens with larger \(g_{j,t}\); and \(\beta\) is the activation exponent, with \(\beta\)=\(0\) giving a purely routing-based criterion, \(\beta\)=\(1\) a norm-like activation score, and \(\beta\)=\(2\) an energy-like score. Under this formulation, many existing one-shot expert-pruning scoring criteria can be viewed as special cases, including Frequency \((0,0,0)\), SEER \((0,1,0)\), EAN \((0,0,1)\), and REAP \((1,1,1)\). It also yields new criteria by exploring alternative choices of the three hyperparameters.

\subsection{Principle for One-shot Expert Pruning}
The formulation provides a systematic basis for selecting expert-pruning scoring criteria under different deployment objectives. We distinguish task-agnostic pruning, which aims to preserve balanced overall performance across heterogeneous capabilities, from task-specific pruning, which aims to preserve or favor a designated capability. To study how each score component relates to these objectives, we use the cross-calibration overlap in \Cref{fig:unified_score_components} as a measure of calibration invariance: if a score identifies nearly the same bottom-ranked experts under different calibration sets, then it is less tied to any single source distribution and is therefore better suited to task-agnostic pruning. As shown in \Cref{fig:unified_score_components}~(\subref{fig:unified_score_components_a}), the dominant effect comes from the routing-frequency term. Switching from \(b\)=\(0\) to \(b\)=\(1\) substantially increases overlap at both pruning ratios, indicating that routing frequency itself is a major source of calibration dependence. For the gate-weight exponent \(\alpha\), smaller values yield higher cross-calibration overlap, with the largest overlap obtained when gate weights are removed, suggesting that gate weights also encode source-specific preferences. By contrast, the role of the activation exponent \(\beta\) is mainly to distinguish routing-only scores from activation-based scores: the main gap is between \(\beta\)=\(0\) and \(\beta\)>\(0\), while the difference between \(\beta\)=\(1\) and \(\beta\)=\(2\) is comparatively small. This leads to a simple principle: task-agnostic pruning should favor gate-free activation scores averaged over routed tokens, while task-specific pruning can benefit from retaining routing-frequency and gate-weight information.

\subsection{Mean (Squared) Activation Norm}
The heatmaps in \Cref{fig:unified_score_components}~(\subref{fig:unified_score_components_b}) instantiate this principle over the full score family. The most calibration-invariant variants concentrate in the \(b=1,\alpha=0\) block, which corresponds exactly to routed-token mean activation scores that remove routing-frequency scaling and discard gate weights. In particular, \((1,0,1)\) and \((1,0,2)\) achieve the highest overlap at both 25\% and 50\% pruning ratios. The model-wise breakdown in Appendix \Cref{fig:appendix_unified_score_model_breakdown} shows the same tendency for the individual models. We denote these variants as Mean Activation Norm (MAN):
\[
\mathrm{MAN}_j
=
\textstyle \frac{1}{N_j}
\sum_{t=1}^{M}
\mathbf{1}[j\in\mathcal{E}_t]\,
\|\mathbf{f}_{j,t}\|_2,
\]
and Mean Squared Activation Norm (MSAN):
\[
\mathrm{MSAN}_j
=
\textstyle \frac{1}{N_j}
\sum_{t=1}^{M}
\mathbf{1}[j\in\mathcal{E}_t]\,
\|\mathbf{f}_{j,t}\|_2^2,
\]
respectively. The score definitions used in this paper are summarized in \Cref{tab:appendix_score_family}.

\section{Experiments}
\subsection{Experimental Setup}
\paragraph{Models and scoring criteria.}
We evaluate on four representative and distinct MoE language models: OLMoE-1B-7B-0125-Instruct~\cite{muennighoff2024olmoe}, DeepSeek-V2-Lite-Chat~\cite{deepseekv2}, ERNIE-4.5-21B-A3B-PT~\cite{baidu2025ernie45}, and Qwen3-30B-A3B-Instruct-2507~\cite{yang2025qwen3}. Together, these models cover a broad range of parameter scales (7B--30B), architectural designs (including the presence or absence of shared experts and dense layers), and active experts per token (top-6 and top-8); architectural details are listed in Appendix \Cref{tab:appendix_model_architecture}. We compare the new scoring criteria derived from our systematic formulation with five representative published baseline criteria: Frequency and SEER~\cite{muzio2024seer}, Expert Activation Norm (EAN)~\cite{jaiswal2025finding}, REAP~\cite{lasby2026reap}, and MoNE~\cite{zhang2026monereplacingredundantexperts}.

\paragraph{Evaluation suite and calibration data.}
For task-agnostic pruning, where the target downstream capability is not assumed known, we use C4~\cite{allenai_c4_2024} as the calibration set, following its common use as a general-domain calibration source in task-agnostic compression~\cite{frantar2023sparsegpt,ling2024slimgpt,xia2024sheared}, and evaluate on 16 downstream benchmarks spanning coding, creative writing, mathematical reasoning, and multiple-choice question answering. For coding, we evaluate on EvalPlus~\cite{liu2023your} and 182 LiveCodeBench~\cite{jain2024livecodebench} problems collected between January and April 2025. For creative writing, we evaluate on 146 prompts sampled from WildBench~\cite{lin2024wildbench}, using \texttt{gpt-oss-120b}~\cite{openai2025gptoss120bgptoss20bmodel} as the judge. For mathematical reasoning, we evaluate on GSM8K~\cite{cobbe2021training} and MATH-500~\cite{hendrycks2021measuring} with EvalScope~\cite{evalscope_2024}. For multiple-choice question answering, we evaluate on MMLU~\cite{hendrycks2020measuring}, AI2 Reasoning Challenge (ARC-C/ARC-E)~\cite{clark2018think}, BoolQ~\cite{clark2019boolq}, OpenBookQA (OBQA)~\cite{mihaylov2018can}, HellaSwag~\cite{zellers2019hellaswag}, Recognizing Textual Entailment (RTE)~\cite{bentivogli2009fifth}, and WinoGrande (WinoG.)~\cite{sakaguchi2021winogrande}, all implemented with lm-eval-harness~\cite{gao2021framework}. For task-specific pruning, we focus on coding and mathematical reasoning, using Evol-CodeAlpaca-v1~\cite{luo2023wizardcoder} and Tulu-3-SFT-Personas-Math~\cite{lambert2024tulu3} as the corresponding capability-aligned calibration sets. In all cases, expert-importance scores are computed from 0.5M tokens sampled from the relevant calibration set.
\paragraph{Protocol and hardware.}
For each model and criterion, we compute expert-importance scores once and uniformly apply the target pruning ratio across MoE layers, removing the lowest-scoring routed experts and their router rows. Shared experts and dense FFN layers, where present, are preserved; routing otherwise follows the original model, without recovery training.
All downstream evaluations are conducted in the zero-shot setting. For open-ended generation benchmarks, we use deterministic decoding with \texttt{do\_sample=False}; when a temperature parameter is exposed, we set \texttt{temperature=0}. This yields greedy, non-sampled generation and improves comparability across pruning methods. All experiments use NVIDIA L40S 48GB GPUs.

\providecommand{\resulttablesep}{{\color{black!55}\vrule width 0.3pt}}
\begin{table*}[t]
  \centering
  \scriptsize
  \setlength{\tabcolsep}{3pt}
  \resizebox{\textwidth}{!}{%
  \begin{tabular}{c c l c!{\resulttablesep}c c!{\resulttablesep}c!{\resulttablesep}c c!{\resulttablesep}c!{\resulttablesep}c c}
    \toprule
    \multicolumn{4}{c!{\resulttablesep}}{} & \multicolumn{2}{c!{\resulttablesep}}{\textbf{Coding}} & \multicolumn{1}{c!{\resulttablesep}}{\textbf{Writing}} & \multicolumn{2}{c!{\resulttablesep}}{\textbf{Math}} & \multicolumn{1}{c!{\resulttablesep}}{\textbf{MC}} & \multicolumn{2}{c}{\textbf{Overall}} \\
    \textbf{Model} & \textbf{Pruning Ratio} & \textbf{Criterion} & \textbf{$(b,\alpha,\beta)$} & \textbf{Eval+} & \textbf{LiveCode} & \textbf{WildBench} & \textbf{GSM8K} & \textbf{MATH-500} & \textbf{MC Avg} & \textbf{Avg} & \textbf{Avg Rank} \\
    \midrule
      \multirow[c]{8}{*}{OLMoE} & 0\% & Full & - & 0.341 & 0.033 & 0.444 & 0.682 & 0.222 & 0.653 & 0.396 & - \\
    \cmidrule(lr){2-12}
       & \multirow[c]{7}{*}{25\%} & Frequency & $(0,0,0)$ & 0.000 & 0.000 & 0.127 & 0.033 & 0.024 & 0.560 & 0.124 & 5.67 \\
       &  & SEER & $(0,1,0)$ & 0.000 & 0.000 & 0.141 & 0.037 & 0.012 & 0.564 & 0.126 & 5.42 \\
       &  & EAN & $(0,0,1)$ & 0.000 & 0.000 & 0.184 & 0.133 & 0.012 & 0.582 & 0.152 & 4.58 \\
       &  & REAP & $(1,1,1)$ & 0.000 & 0.000 & \textbf{0.263} & 0.139 & 0.036 & \textbf{0.601} & 0.173 & 2.83 \\
       &  & MoNE & - & 0.000 & 0.000 & 0.181 & 0.117 & 0.006 & 0.583 & 0.148 & 5.00 \\
    \cdashline{3-12}
       &  & \cellcolor{bluerow}MAN & \cellcolor{bluerow}$(1,0,1)$ & \cellcolor{bluerow}\textbf{0.009} & \cellcolor{bluerow}0.000 & \cellcolor{bluerow}\underline{0.260} & \cellcolor{bluerow}\textbf{0.208} & \cellcolor{bluerow}\underline{0.046} & \cellcolor{bluerow}\underline{0.594} & \cellcolor{bluerow}\textbf{0.186} & \cellcolor{bluerow}\textbf{2.00} \\
       &  & \cellcolor{bluerow}MSAN & \cellcolor{bluerow}$(1,0,2)$ & \cellcolor{bluerow}\underline{0.008} & \cellcolor{bluerow}0.000 & \cellcolor{bluerow}0.242 & \cellcolor{bluerow}\underline{0.194} & \cellcolor{bluerow}\textbf{0.056} & \cellcolor{bluerow}0.589 & \cellcolor{bluerow}\underline{0.181} & \cellcolor{bluerow}\underline{2.50} \\
    \midrule[\heavyrulewidth]
      \multirow[c]{8}{*}{DeepSeek} & 0\% & Full & - & 0.549 & 0.104 & 0.418 & 0.610 & 0.298 & 0.678 & 0.443 & - \\
    \cmidrule(lr){2-12}
       & \multirow[c]{7}{*}{25\%} & Frequency & $(0,0,0)$ & 0.000 & 0.000 & \underline{0.291} & 0.023 & 0.012 & 0.602 & 0.155 & 5.33 \\
       &  & SEER & $(0,1,0)$ & 0.000 & 0.000 & 0.155 & 0.034 & 0.016 & 0.602 & 0.134 & 5.67 \\
       &  & EAN & $(0,0,1)$ & 0.000 & 0.000 & \textbf{0.295} & \underline{0.312} & 0.028 & 0.621 & \underline{0.209} & \underline{3.33} \\
       &  & REAP & $(1,1,1)$ & \underline{0.007} & 0.000 & 0.174 & 0.281 & 0.028 & 0.620 & 0.185 & 3.92 \\
       &  & MoNE & - & 0.000 & 0.000 & 0.200 & 0.227 & 0.024 & 0.629 & 0.180 & 4.42 \\
    \cdashline{3-12}
       &  & \cellcolor{bluerow}MAN & \cellcolor{bluerow}$(1,0,1)$ & \cellcolor{bluerow}0.001 & \cellcolor{bluerow}0.000 & \cellcolor{bluerow}0.154 & \cellcolor{bluerow}0.287 & \cellcolor{bluerow}\underline{0.032} & \cellcolor{bluerow}\textbf{0.636} & \cellcolor{bluerow}0.185 & \cellcolor{bluerow}\underline{3.33} \\
       &  & \cellcolor{bluerow}MSAN & \cellcolor{bluerow}$(1,0,2)$ & \cellcolor{bluerow}\textbf{0.025} & \cellcolor{bluerow}0.000 & \cellcolor{bluerow}0.238 & \cellcolor{bluerow}\textbf{0.428} & \cellcolor{bluerow}\textbf{0.102} & \cellcolor{bluerow}\underline{0.630} & \cellcolor{bluerow}\textbf{0.237} & \cellcolor{bluerow}\textbf{2.00} \\
    \midrule[\heavyrulewidth]
      \multirow[c]{8}{*}{ERNIE} & 0\% & Full & - & 0.867 & 0.247 & 0.479 & 0.829 & 0.780 & 0.721 & 0.654 & - \\
    \cmidrule(lr){2-12}
       & \multirow[c]{7}{*}{25\%} & Frequency & $(0,0,0)$ & 0.254 & 0.055 & 0.352 & 0.647 & 0.316 & 0.655 & 0.380 & 6.67 \\
       &  & SEER & $(0,1,0)$ & 0.256 & 0.060 & 0.381 & 0.748 & 0.368 & 0.658 & 0.412 & 5.25 \\
       &  & EAN & $(0,0,1)$ & 0.300 & 0.055 & 0.408 & 0.673 & 0.370 & 0.682 & 0.415 & 4.50 \\
       &  & REAP & $(1,1,1)$ & 0.277 & 0.060 & \underline{0.414} & 0.760 & 0.528 & 0.700 & 0.456 & \underline{3.25} \\
       &  & MoNE & - & 0.217 & 0.055 & \textbf{0.419} & 0.774 & 0.384 & 0.678 & 0.421 & 4.33 \\
    \cdashline{3-12}
       &  & \cellcolor{bluerow}MAN & \cellcolor{bluerow}$(1,0,1)$ & \cellcolor{bluerow}\underline{0.343} & \cellcolor{bluerow}\textbf{0.077} & \cellcolor{bluerow}0.402 & \cellcolor{bluerow}\underline{0.801} & \cellcolor{bluerow}\textbf{0.580} & \cellcolor{bluerow}\textbf{0.705} & \cellcolor{bluerow}\textbf{0.485} & \cellcolor{bluerow}\textbf{2.00} \\
       &  & \cellcolor{bluerow}MSAN & \cellcolor{bluerow}$(1,0,2)$ & \cellcolor{bluerow}\textbf{0.354} & \cellcolor{bluerow}\underline{0.066} & \cellcolor{bluerow}0.404 & \cellcolor{bluerow}\textbf{0.813} & \cellcolor{bluerow}\underline{0.564} & \cellcolor{bluerow}\underline{0.704} & \cellcolor{bluerow}\underline{0.484} & \cellcolor{bluerow}\textbf{2.00} \\
    \midrule[\heavyrulewidth]
      \multirow[c]{8}{*}{Qwen3} & 0\% & Full & - & 0.871 & 0.368 & 0.644 & 0.923 & 0.802 & 0.737 & 0.724 & - \\
    \cmidrule(lr){2-12}
       & \multirow[c]{7}{*}{25\%} & Frequency & $(0,0,0)$ & 0.000 & 0.000 & \textbf{0.632} & 0.904 & 0.196 & 0.732 & 0.411 & 4.83 \\
       &  & SEER & $(0,1,0)$ & 0.003 & 0.000 & 0.612 & 0.913 & 0.202 & \underline{0.734} & 0.411 & 3.92 \\
       &  & EAN & $(0,0,1)$ & 0.001 & 0.000 & 0.623 & 0.910 & 0.194 & \textbf{0.735} & 0.411 & 4.42 \\
       &  & REAP & $(1,1,1)$ & 0.599 & 0.137 & 0.600 & 0.879 & 0.778 & 0.721 & 0.619 & 4.67 \\
       &  & MoNE & - & 0.000 & 0.000 & \underline{0.627} & 0.911 & 0.206 & 0.733 & 0.413 & 4.17 \\
    \cdashline{3-12}
       &  & \cellcolor{bluerow}MAN & \cellcolor{bluerow}$(1,0,1)$ & \cellcolor{bluerow}\textbf{0.868} & \cellcolor{bluerow}\textbf{0.346} & \cellcolor{bluerow}0.570 & \cellcolor{bluerow}\textbf{0.937} & \cellcolor{bluerow}\underline{0.792} & \cellcolor{bluerow}0.727 & \cellcolor{bluerow}\textbf{0.707} & \cellcolor{bluerow}\textbf{2.75} \\
       &  & \cellcolor{bluerow}MSAN & \cellcolor{bluerow}$(1,0,2)$ & \cellcolor{bluerow}\underline{0.847} & \cellcolor{bluerow}\underline{0.335} & \cellcolor{bluerow}0.559 & \cellcolor{bluerow}\underline{0.935} & \cellcolor{bluerow}\textbf{0.796} & \cellcolor{bluerow}0.727 & \cellcolor{bluerow}\underline{0.700} & \cellcolor{bluerow}\underline{3.25} \\
    \bottomrule
  \end{tabular}%
  }
  \caption{\textbf{Task-agnostic pruning with C4 calibration.} We report results at a 25\% pruning ratio for four representative MoE models, comparing published baseline scoring criteria with the proposed mean-activation criteria MAN and MSAN (blue). Benchmarks are grouped by capability; Avg and Avg Rank summarize balanced performance across the displayed benchmark columns. Bold and underline indicate the best and second-best results within each block.}
  \label{tab:pvar_task_agnostic_c4}
\end{table*}

\providecommand{\resulttablesep}{{\color{black!55}\vrule width 0.3pt}}
\begin{table*}[t]
  \centering
  \small
  \setlength{\tabcolsep}{3pt}
  \resizebox{\textwidth}{!}{%
  \begin{tabular}{c c l!{\resulttablesep} c c!{\resulttablesep} c c!{\resulttablesep} c!{\resulttablesep} c!{\resulttablesep} c c!{\resulttablesep} c c!{\resulttablesep} c}
    \toprule
    \multicolumn{3}{c!{\resulttablesep}}{} & \multicolumn{2}{c!{\resulttablesep}}{\textbf{Coding}} & \multicolumn{2}{c!{\resulttablesep}}{\textbf{Math}} & \multicolumn{1}{c!{\resulttablesep}}{\textbf{Rank}} & \multicolumn{1}{c!{\resulttablesep}}{} & \multicolumn{2}{c!{\resulttablesep}}{\textbf{Coding}} & \multicolumn{2}{c!{\resulttablesep}}{\textbf{Math}} & \multicolumn{1}{c}{\textbf{Rank}} \\
    \textbf{Model} & \textbf{Pruning Ratio} & \textbf{Criterion} & \textbf{Eval+} & \textbf{LiveCode} & \textbf{GSM8K} & \textbf{MATH-500} & \textbf{Avg Rank} & \textbf{Model} & \textbf{Eval+} & \textbf{LiveCode} & \textbf{GSM8K} & \textbf{MATH-500} & \textbf{Avg Rank} \\
    \midrule
      \multirow[c]{10}{*}{OLMoE} & 0\% & Full & 0.341 & 0.033 & 0.682 & 0.222 & - & \multirow[c]{10}{*}{DeepSeek} & 0.549 & 0.104 & 0.610 & 0.298 & - \\
    \cmidrule(lr){2-8}\cmidrule(lr){10-14}
       & \multirow[c]{9}{*}{25\%} & Frequency & 0.341 & \underline{0.022} & 0.593 & \underline{0.226} & 3.75 &  & 0.297 & \textbf{0.099} & 0.476 & 0.188 & 6.62 \\
       &  & SEER & 0.339 & \textbf{0.027} & 0.594 & 0.194 & 4.88 &  & 0.404 & \textbf{0.099} & 0.477 & 0.196 & 5.75 \\
       &  & EAN & \underline{0.345} & 0.016 & \underline{0.622} & 0.210 & \underline{3.50} &  & 0.440 & \underline{0.088} & 0.479 & 0.172 & 6.00 \\
       &  & REAP & 0.300 & 0.011 & \textbf{0.626} & 0.198 & 5.62 &  & \textbf{0.465} & \underline{0.088} & \textbf{0.585} & \underline{0.218} & \textbf{2.00} \\
       &  & MoNE & 0.339 & 0.011 & 0.618 & 0.198 & 5.75 &  & 0.406 & \underline{0.088} & \underline{0.581} & 0.210 & 4.25 \\
    \cdashline{3-8}\cdashline{10-14}
    \addlinespace[1pt]
       &  & MAN & 0.243 & 0.016 & 0.541 & 0.216 & 6.62 &  & \underline{0.444} & 0.077 & 0.551 & 0.196 & 5.12 \\
       &  & MSAN & 0.254 & \textbf{0.027} & 0.558 & 0.180 & 6.75 &  & 0.411 & 0.082 & 0.561 & \textbf{0.226} & 4.38 \\
       &  & \cellcolor{bluerow}RCN & \cellcolor{bluerow}0.333 & \cellcolor{bluerow}\textbf{0.027} & \cellcolor{bluerow}0.619 & \cellcolor{bluerow}\textbf{0.246} & \cellcolor{bluerow}\textbf{3.00} &  & \cellcolor{bluerow}0.435 & \cellcolor{bluerow}0.066 & \cellcolor{bluerow}0.500 & \cellcolor{bluerow}0.186 & \cellcolor{bluerow}7.00 \\
       &  & \cellcolor{bluerow}RCE & \cellcolor{bluerow}\textbf{0.348} & \cellcolor{bluerow}0.011 & \cellcolor{bluerow}0.581 & \cellcolor{bluerow}0.210 & \cellcolor{bluerow}5.12 &  & \cellcolor{bluerow}0.442 & \cellcolor{bluerow}0.082 & \cellcolor{bluerow}0.578 & \cellcolor{bluerow}0.216 & \cellcolor{bluerow}\underline{3.88} \\
    \midrule[\heavyrulewidth]
      \multirow[c]{19}{*}{ERNIE} & 0\% & Full & 0.867 & 0.247 & 0.829 & 0.780 & - & \multirow[c]{19}{*}{Qwen3} & 0.871 & 0.368 & 0.923 & 0.802 & - \\
    \cmidrule(lr){2-8}\cmidrule(lr){10-14}
       & \multirow[c]{9}{*}{25\%} & Frequency & 0.818 & 0.181 & \textbf{0.832} & 0.736 & 7.25 &  & 0.862 & 0.357 & 0.897 & \underline{0.794} & 6.00 \\
       &  & SEER & 0.830 & 0.214 & 0.822 & 0.768 & 6.50 &  & 0.851 & 0.363 & 0.898 & \textbf{0.806} & 6.00 \\
       &  & EAN & 0.821 & \textbf{0.231} & 0.825 & 0.774 & 5.75 &  & 0.864 & \textbf{0.396} & \textbf{0.917} & 0.792 & \textbf{2.62} \\
       &  & REAP & \underline{0.835} & \textbf{0.231} & 0.829 & 0.792 & \underline{3.38} &  & 0.852 & 0.335 & 0.904 & \textbf{0.806} & 6.25 \\
       &  & MoNE & 0.826 & 0.209 & 0.830 & \textbf{0.800} & 5.12 &  & 0.859 & 0.363 & \underline{0.914} & 0.780 & 5.62 \\
    \cdashline{3-8}\cdashline{10-14}
    \addlinespace[1pt]
       &  & MAN & 0.830 & 0.214 & \textbf{0.832} & 0.776 & 4.25 &  & \textbf{0.871} & \textbf{0.396} & 0.904 & 0.788 & 4.00 \\
       &  & MSAN & \underline{0.835} & 0.209 & \textbf{0.832} & 0.748 & 5.00 &  & 0.860 & 0.363 & 0.905 & 0.776 & 6.25 \\
       &  & \cellcolor{bluerow}RCN & \cellcolor{bluerow}0.834 & \cellcolor{bluerow}\textbf{0.231} & \cellcolor{bluerow}0.827 & \cellcolor{bluerow}0.760 & \cellcolor{bluerow}5.00 &  & \cellcolor{bluerow}\underline{0.868} & \cellcolor{bluerow}\underline{0.379} & \cellcolor{bluerow}0.912 & \cellcolor{bluerow}0.792 & \cellcolor{bluerow}\underline{3.38} \\
       &  & \cellcolor{bluerow}RCE & \cellcolor{bluerow}\textbf{0.837} & \cellcolor{bluerow}\underline{0.225} & \cellcolor{bluerow}\underline{0.831} & \cellcolor{bluerow}\underline{0.796} & \cellcolor{bluerow}\textbf{2.75} &  & \cellcolor{bluerow}0.860 & \cellcolor{bluerow}0.363 & \cellcolor{bluerow}0.912 & \cellcolor{bluerow}0.792 & \cellcolor{bluerow}4.88 \\
    \cmidrule(lr){2-8}\cmidrule(lr){10-14}
       & \multirow[c]{9}{*}{50\%} & Frequency & 0.647 & 0.143 & 0.729 & 0.554 & 9.00 &  & 0.700 & 0.225 & 0.858 & 0.764 & 7.88 \\
       &  & SEER & 0.698 & 0.170 & 0.766 & 0.634 & 6.00 &  & 0.700 & 0.247 & 0.857 & 0.766 & 7.62 \\
       &  & EAN & 0.657 & 0.148 & \underline{0.801} & 0.622 & 6.00 &  & 0.841 & 0.313 & 0.874 & \textbf{0.800} & \underline{4.25} \\
       &  & REAP & \underline{0.737} & 0.198 & 0.765 & \textbf{0.680} & 3.50 &  & 0.835 & \textbf{0.357} & \underline{0.884} & 0.740 & 4.50 \\
       &  & MoNE & \textbf{0.741} & 0.176 & 0.785 & 0.616 & 4.75 &  & 0.842 & 0.346 & 0.872 & 0.780 & 4.62 \\
    \cdashline{3-8}\cdashline{10-14}
    \addlinespace[1pt]
       &  & MAN & 0.716 & \textbf{0.214} & 0.790 & \underline{0.676} & \textbf{2.62} &  & 0.850 & 0.346 & 0.861 & 0.750 & 5.38 \\
       &  & MSAN & 0.692 & 0.176 & 0.785 & 0.632 & 5.75 &  & 0.832 & 0.324 & 0.856 & 0.782 & 6.50 \\
       &  & \cellcolor{bluerow}RCN & \cellcolor{bluerow}0.706 & \cellcolor{bluerow}\underline{0.203} & \cellcolor{bluerow}0.790 & \cellcolor{bluerow}0.604 & \cellcolor{bluerow}4.62 &  & \cellcolor{bluerow}\textbf{0.855} & \cellcolor{bluerow}\underline{0.352} & \cellcolor{bluerow}0.883 & \cellcolor{bluerow}\underline{0.794} & \cellcolor{bluerow}\textbf{2.12} \\
       &  & \cellcolor{bluerow}RCE & \cellcolor{bluerow}0.733 & \cellcolor{bluerow}0.192 & \cellcolor{bluerow}\textbf{0.804} & \cellcolor{bluerow}0.654 & \cellcolor{bluerow}\underline{2.75} &  & \cellcolor{bluerow}\underline{0.854} & \cellcolor{bluerow}\underline{0.352} & \cellcolor{bluerow}\textbf{0.892} & \cellcolor{bluerow}0.784 & \cellcolor{bluerow}\textbf{2.12} \\
    \bottomrule
  \end{tabular}%
  }
  \caption{\textbf{Task-specific pruning with capability-aligned calibration.}
For coding, expert-importance scores are computed on Evol-CodeAlpaca-v1; for math, scores are computed on Tulu-3-SFT-Personas-Math. We compare published baseline scoring criteria and MAN/MSAN with routing-frequency-retaining and gate-weighted criteria from our systematic scoring formulation, which are in principle suited to task-specific pruning (blue).}
  \label{tab:pvar_task_specific_code_math_with_msan}
\end{table*}

\begin{table}[!t]
  \centering
  \small
  \resizebox{\columnwidth}{!}{%
  \begin{tabular}{l c c}
    \toprule
    \textbf{Model} & \textbf{GPU} & \textbf{Loading Memory (GB)} \\
    \midrule
    OLMoE-7B & 1$\times$L40S & 6.89 / 12.89 \\
    DeepSeek-16B & 1$\times$L40S & 15.86 / 29.27 \\
    ERNIE-21B & 2$\times$L40S & 21.67 / 40.66 \\
    Qwen3-30B & 2$\times$L40S & 29.93 / 56.92 \\
    \bottomrule
  \end{tabular}%
  }
  \caption{\textbf{Model loading memory before and after pruning.} Models are loaded in bf16. GPU denotes the number of L40S GPUs needed to load the model and perform one-shot pruning. Loading memory is reported as after / before pruning at 50\% pruning ratio.}
  \label{tab:efficiency_comparison}
  \vspace{-2em}
\end{table}

\subsection{Results}
\noindent\textbf{Task-agnostic results.}
\Cref{tab:pvar_task_agnostic_c4} evaluates one-shot pruning under the task-agnostic setting, where expert-importance scores are computed on C4 and the pruned models are evaluated across coding, writing, math, and multiple-choice benchmarks; the full per-benchmark results, including the 50\% pruning ratio, are reported in \Cref{tab:pvar_task_agnostic_c4_appendix,tab:pvar_task_agnostic_c4_appendix_mc}. The main trend is clear: the routed-token-averaged, gate-free mean-activation scores, MAN \((1,0,1)\) and MSAN \((1,0,2)\), are the strongest overall choices. Across all four representative MoE models, either MAN or MSAN achieves the best overall average and the best average rank. Relative to the strongest prior baseline in each model, the best mean-activation criterion improves the overall average by \(+1.3\), \(+2.8\), \(+2.9\), and \(+8.8\) percentage points on OLMoE, DeepSeek, ERNIE, and Qwen3, respectively. These gains are obtained without task-specific calibration or recovery finetuning. The per-capability results further clarify why MAN and MSAN are preferable for task-agnostic pruning. Their advantage is not merely an artifact of one benchmark group: on ERNIE, MAN/MSAN improve the overall average while also giving the strongest or near-strongest results on coding, math, and multiple-choice evaluation. On Qwen3, the gap is particularly large for coding: under C4 calibration, Frequency, SEER, EAN, and MoNE nearly collapse on EvalPlus and LiveCodeBench, whereas MAN preserves coding performance close to the full model. Appendix~\ref{sec:appendix_task_agnostic_robustness} further shows substantial coding and math recovery for OLMoE and DeepSeek at 12.5\% pruning, with MAN or MSAN achieving the highest overall average at this ratio and under code-domain calibration. Using routed-token averages and removing gate weighting makes MAN/MSAN more reliable across diverse downstream evaluations, which explains their consistently stronger average performance and ranking.

\noindent\textbf{Task-specific pruning with capability-aligned calibration.}
\Cref{tab:pvar_task_specific_code_math_with_msan} evaluates task-specific pruning, where expert-importance scores for coding are computed on Evol-CodeAlpaca-v1 and scores for math are computed on Tulu-3-SFT-Personas-Math. Unlike the task-agnostic setting, the calibration distribution is intentionally aligned with the target capability. In this case, routing-frequency and gate-weight signals become informative rather than nuisances, and the strongest scoring criteria tend to combine all three factors in our formulation: routing frequency, gate magnitude, and activation strength. This trend is visible when comparing MAN/MSAN with the routing-frequency-retaining, gate-weighted activation criteria. MAN and MSAN use routed-token averages and remove gate weighting, which makes them strong for task-agnostic pruning but less consistently optimal here. By contrast, Routed Contribution Norm (RCN; \(0,1,1\)) and Routed Contribution Energy (RCE; \(0,2,2\)) derived from our formulation retain routing frequency, gate weighting, and activation information. They achieve the best average rank on OLMoE at a 25\% pruning ratio, ERNIE at a 25\% pruning ratio, and Qwen3 at a 50\% pruning ratio, while remaining competitive in most other settings. The main exceptions are DeepSeek, where REAP is strongest, and Qwen3 at a 25\% pruning ratio, where EAN achieves the best average rank. Overall, the capability-aligned calibration results support the task-specific side of our principle: when pruning for a known target capability, expert importance should reflect not only how strongly an expert activates, but also how often and how confidently the router uses it on matched data. Thus, task-specific pruning benefits from retaining routing-frequency, gate-weight, and activation signals jointly, whereas MAN/MSAN are better suited to the task-agnostic setting where routing effects should be suppressed.

\noindent\textbf{Loading memory.}
\Cref{tab:efficiency_comparison} shows that at a 50\% pruning ratio, expert pruning reduces loading memory by \(\sim\)50\% across the evaluated MoE models, validating it as a practical way to address memory usage, the main bottleneck in MoE deployment.
\FloatBarrier

\section{Conclusion}
We focus on selecting an expert-importance criterion suited to a given deployment objective for one-shot MoE expert pruning, rather than identifying a universally dominant score. From a single-expert removal analysis, we derived a systematic formulation that decomposes scoring criteria into routing frequency, gate weighting, and activation strength. This formulation exposes a criteria selection principle: task-agnostic pruning should suppress calibration-specific routing effects by using routed-token-averaged, gate-free activation criteria, whereas task-specific pruning can benefit from routing-frequency and gate-weight signals when calibration data are aligned with the target capability.
This perspective explains why routing-only criteria such as Frequency and SEER are sensitive to calibration distributions, and why activation-based criteria can be adapted to different objectives through explicit choices about routed-token averaging and gate weighting. It also yields MAN and MSAN, two new task-agnostic criteria that better preserve balanced performance across heterogeneous benchmarks, improving the average by up to 8.8 percentage points over the strongest prior baseline. Overall, our results provide a compact and empirically supported basis for choosing one-shot expert-pruning criteria under different deployment requirements while keeping the resulting method simple and broadly applicable in practice.

\section*{Limitations}
Our systematic formulation covers commonly used one-shot expert-pruning criteria including Frequency, SEER, EAN, and REAP. Although the principle suggested by this formulation---that routing-related signals introduce calibration dependence and should therefore be suppressed for task-agnostic pruning while retained when the calibration data are target-aligned---may extend beyond this exact score family, not all criteria can be represented in the same form. For example, MoNE uses routing-frequency-weighted activation variance across routed tokens to characterize expert redundancy. Although its routing-frequency component is related to our formulation and the resulting principle may partially transfer, the variance statistic itself is not fully captured by \(\mathcal{S}(b,\alpha,\beta)\). In addition, our study focuses on expert pruning, where experts are removed according to importance scores; expert merging modifies multiple experts jointly and may require a different formulation and selection principle.

\section*{Acknowledgements}

This work was supported by National Key R\&D Program of China
(2024YFF0505603), the National Natural Science Foundation of China
(grant number U2541205, 62271414), Zhejiang Provincial Distinguished Young
Scientist Foundation (grant number LR23F010001), Zhejiang `Pioneer' and `Leading Goose'
R\&D Program (grant number 2024SDXHDX0006, 2024C03182), the Key Project of
Westlake Institute for Optoelectronics (grant number 2023GD007), and the 2023
International Sci-tech Cooperation Projects under the purview of the
`Innovation Yongjiang 2035' Key R\&D Program (grant number 2024Z126).

\bibliography{custom}

\clearpage
\appendix

\twocolumn[{%
\begin{minipage}{\textwidth}
\captionsetup{hypcap=false}
\section{Model Architecture Details}
\label{sec:appendix_model_architecture}
{  \centering
  \small
  \resizebox{\textwidth}{!}{%
  \begin{tabular}{l c c c c l}
    \toprule
    \textbf{Model} & \textbf{Total params} & \textbf{Experts per layer} & \textbf{Active experts per token} & \textbf{Shared experts} & \textbf{Dense/MoE layout} \\
    \midrule
    OLMoE-1B-7B-0125-Instruct & 7B & 64 & top-8 & No & 0 dense, 16 MoE layers \\
    DeepSeek-V2-Lite-Chat & 16B & 64 routed & top-6 & Yes (2 shared) & 1 dense, 26 MoE layers \\
    ERNIE-4.5-21B-A3B-PT & 21B & 64 routed & top-6 & Yes (2 shared) & 1 dense, 27 MoE layers \\
    Qwen3-30B-A3B-Instruct-2507 & 30B & 128 & top-8 & No & 0 dense, 48 MoE layers \\
    \bottomrule
  \end{tabular}%
  }
  \captionof{table}{\textbf{Architectural diversity of the evaluated MoE models.}}
  \label{tab:appendix_model_architecture}
}

\section{Model-wise Calibration Overlap}
\label{sec:appendix_modelwise_overlap}
{\centering
  \includegraphics[width=0.82\textwidth]{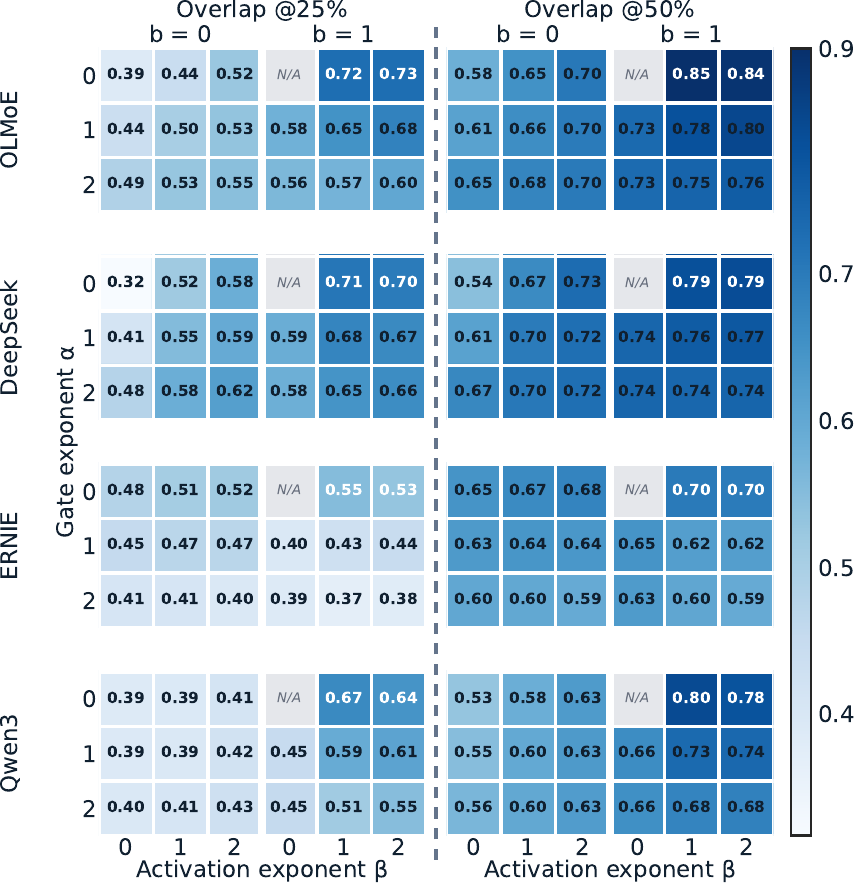}
  \captionof{figure}{\textbf{Model-wise overlap of bottom-ranked experts across calibration distributions.} Each cell reports the overlap among bottom-ranked experts selected by a score variant \(S_j(b,\alpha,\beta)\) when rankings are computed from C4, Evol-CodeAlpaca-v1, and Tulu-3-SFT-Personas-Math. Results are shown separately for each model at 25\% and 50\% pruning ratios. N/A denotes the degenerate case \((1,0,0)\). The averaged summary is shown in \Cref{fig:unified_score_components}.}
  \label{fig:appendix_unified_score_model_breakdown}\par}
\end{minipage}
}]
\clearpage

\twocolumn[{%
\begin{minipage}{\textwidth}
\captionsetup{hypcap=false}
\begin{minipage}[t]{0.48\textwidth}
\vspace{0pt}
\section{Additional Task-Agnostic Robustness Results}
\label{sec:appendix_task_agnostic_robustness}

\noindent\textbf{Robustness to pruning severity.}
To examine whether the substantial degradation in coding and math performance at 25\% is driven by pruning severity, we additionally evaluate OLMoE and DeepSeek at 12.5\%. The results show substantial recovery in both categories and the highest overall averages for MAN/MSAN on both models (\Cref{tab:appendix_task_agnostic_c4_0125}), indicating that the degradation at 25\% is tied to pruning severity.
\end{minipage}
\hfill
\begin{minipage}[t]{0.48\textwidth}
\vspace{0pt}
\noindent\textbf{Robustness to the calibration distribution.}
To test calibration sensitivity, we replace C4 with Evol-CodeAlpaca-v1 and evaluate on the same 16 benchmarks. MAN or MSAN attains the highest overall average in every model--ratio setting, and together they outperform REAP in 9 of 10 direct comparisons; the sole exception, MSAN on ERNIE at 50\%, trails by 0.001 (\Cref{tab:appendix_task_agnostic_evol_codealpaca}). REAP often leads on coding under this code-domain calibration, whereas MAN/MSAN remain more balanced across tasks, supporting our task-agnostic selection principle.
\end{minipage}

\vspace{0.5em}
{\providecommand{\resulttablesep}{{\color{black!55}\vrule width 0.3pt}}
  \centering
  \scriptsize
  \setlength{\tabcolsep}{3pt}
  \resizebox{\textwidth}{!}{%
  \begin{tabular}{c c l c!{\resulttablesep}c c!{\resulttablesep}c!{\resulttablesep}c c!{\resulttablesep}c!{\resulttablesep}c c}
    \toprule
    \multicolumn{4}{c!{\resulttablesep}}{} & \multicolumn{2}{c!{\resulttablesep}}{\textbf{Coding}} & \multicolumn{1}{c!{\resulttablesep}}{\textbf{Writing}} & \multicolumn{2}{c!{\resulttablesep}}{\textbf{Math}} & \multicolumn{1}{c!{\resulttablesep}}{\textbf{MC}} & \multicolumn{2}{c}{\textbf{Overall}} \\
    \textbf{Model} & \textbf{Pruning Ratio} & \textbf{Criterion} & \textbf{$(b,\alpha,\beta)$} & \textbf{Eval+} & \textbf{LiveCode} & \textbf{WildBench} & \textbf{GSM8K} & \textbf{MATH-500} & \textbf{MC Avg} & \textbf{Avg} & \textbf{Avg Rank} \\
    \midrule
      \multirow[c]{8}{*}{OLMoE} & 0\% & Full & - & 0.341 & 0.033 & 0.444 & 0.682 & 0.222 & 0.653 & 0.396 & - \\
    \cmidrule(lr){2-12}
       & \multirow[c]{7}{*}{12.5\%} & Frequency & $(0,0,0)$ & 0.000 & 0.000 & \underline{0.368} & 0.282 & 0.024 & 0.634 & 0.218 & 4.92 \\
       &  & SEER & $(0,1,0)$ & 0.000 & 0.000 & 0.356 & 0.292 & 0.018 & 0.630 & 0.216 & 5.92 \\
       &  & EAN & $(0,0,1)$ & 0.000 & 0.000 & \textbf{0.373} & 0.354 & 0.034 & \underline{0.643} & 0.234 & 3.58 \\
       &  & REAP & $(1,1,1)$ & 0.020 & 0.000 & 0.364 & 0.607 & 0.128 & 0.641 & 0.293 & \underline{3.42} \\
       &  & MoNE & - & 0.000 & 0.000 & 0.364 & 0.331 & 0.038 & \textbf{0.646} & 0.230 & 3.92 \\
    \cdashline{3-12}
       &  & \cellcolor{bluerow}MAN & \cellcolor{bluerow}$(1,0,1)$ & \cellcolor{bluerow}\textbf{0.075} & \cellcolor{bluerow}0.000 & \cellcolor{bluerow}0.362 & \cellcolor{bluerow}\textbf{0.650} & \cellcolor{bluerow}\underline{0.184} & \cellcolor{bluerow}0.628 & \cellcolor{bluerow}\textbf{0.317} & \cellcolor{bluerow}3.50 \\
       &  & \cellcolor{bluerow}MSAN & \cellcolor{bluerow}$(1,0,2)$ & \cellcolor{bluerow}\underline{0.062} & \cellcolor{bluerow}0.000 & \cellcolor{bluerow}0.364 & \cellcolor{bluerow}\underline{0.644} & \cellcolor{bluerow}\textbf{0.188} & \cellcolor{bluerow}0.641 & \cellcolor{bluerow}\textbf{0.317} & \cellcolor{bluerow}\textbf{2.75} \\
    \midrule[\heavyrulewidth]
      \multirow[c]{8}{*}{DeepSeek} & 0\% & Full & - & 0.549 & 0.104 & 0.418 & 0.610 & 0.298 & 0.678 & 0.443 & - \\
    \cmidrule(lr){2-12}
       & \multirow[c]{7}{*}{12.5\%} & Frequency & $(0,0,0)$ & 0.000 & 0.000 & \underline{0.377} & 0.068 & 0.026 & 0.656 & 0.188 & 5.67 \\
       &  & SEER & $(0,1,0)$ & 0.001 & 0.000 & 0.355 & 0.412 & 0.062 & 0.655 & 0.247 & 5.50 \\
       &  & EAN & $(0,0,1)$ & 0.010 & 0.000 & 0.365 & 0.446 & 0.050 & 0.663 & 0.256 & 4.83 \\
       &  & REAP & $(1,1,1)$ & 0.046 & 0.000 & 0.222 & 0.410 & 0.088 & \textbf{0.673} & 0.240 & 4.17 \\
       &  & MoNE & - & 0.012 & 0.000 & \textbf{0.396} & 0.475 & 0.076 & 0.667 & 0.271 & 3.33 \\
    \cdashline{3-12}
       &  & \cellcolor{bluerow}MAN & \cellcolor{bluerow}$(1,0,1)$ & \cellcolor{bluerow}\textbf{0.289} & \cellcolor{bluerow}\textbf{0.016} & \cellcolor{bluerow}0.280 & \cellcolor{bluerow}\underline{0.579} & \cellcolor{bluerow}\underline{0.178} & \cellcolor{bluerow}0.664 & \cellcolor{bluerow}\underline{0.334} & \cellcolor{bluerow}\underline{2.67} \\
       &  & \cellcolor{bluerow}MSAN & \cellcolor{bluerow}$(1,0,2)$ & \cellcolor{bluerow}\underline{0.223} & \cellcolor{bluerow}\underline{0.005} & \cellcolor{bluerow}0.368 & \cellcolor{bluerow}\textbf{0.582} & \cellcolor{bluerow}\textbf{0.188} & \cellcolor{bluerow}\underline{0.671} & \cellcolor{bluerow}\textbf{0.339} & \cellcolor{bluerow}\textbf{1.83} \\
    \bottomrule
  \end{tabular}%
  }
  \captionof{table}{\textbf{Task-agnostic pruning at a lower pruning ratio.} Results for OLMoE and DeepSeek under C4 calibration at 12.5\% pruning. Eval+ and MC Avg aggregate four EvalPlus and eight multiple-choice benchmarks; Avg averages the six displayed benchmark columns. Bold and underline denote the best and second-best criterion within each model block.}
  \label{tab:appendix_task_agnostic_c4_0125}
}
\vspace{0.5em}
{\providecommand{\resulttablesep}{{\color{black!55}\vrule width 0.3pt}}
  \centering
  \scriptsize
  \setlength{\tabcolsep}{4pt}
  \noindent\makebox[\textwidth][c]{%
  \resizebox{0.9\textwidth}{!}{%
  \begin{tabular}{c c l!{\resulttablesep}c c!{\resulttablesep}c!{\resulttablesep}c c!{\resulttablesep}c!{\resulttablesep}c}
    \toprule
    \multicolumn{3}{c!{\resulttablesep}}{} & \multicolumn{2}{c!{\resulttablesep}}{\textbf{Coding}} & \multicolumn{1}{c!{\resulttablesep}}{\textbf{Writing}} & \multicolumn{2}{c!{\resulttablesep}}{\textbf{Math}} & \multicolumn{1}{c!{\resulttablesep}}{\textbf{MC}} & \multicolumn{1}{c}{\textbf{Overall}} \\
    \textbf{Model} & \textbf{Pruning Ratio} & \textbf{Criterion} & \textbf{Eval+} & \textbf{LiveCode} & \textbf{WildBench} & \textbf{GSM8K} & \textbf{MATH-500} & \textbf{MC Avg} & \textbf{Avg} \\
    \midrule
      \multirow[c]{3}{*}{DeepSeek} & \multirow[c]{3}{*}{25\%} & REAP & \textbf{0.465} & \textbf{0.088} & 0.261 & 0.546 & \underline{0.216} & 0.610 & 0.364 \\
    \cdashline{3-10}
       & & \cellcolor{bluerow}MAN & \cellcolor{bluerow}\underline{0.444} & \cellcolor{bluerow}0.077 & \cellcolor{bluerow}\textbf{0.313} & \cellcolor{bluerow}\underline{0.566} & \cellcolor{bluerow}\textbf{0.224} & \cellcolor{bluerow}\underline{0.617} & \cellcolor{bluerow}\textbf{0.373} \\
       & & \cellcolor{bluerow}MSAN & \cellcolor{bluerow}0.411 & \cellcolor{bluerow}\underline{0.082} & \cellcolor{bluerow}\underline{0.307} & \cellcolor{bluerow}\textbf{0.575} & \cellcolor{bluerow}0.212 & \cellcolor{bluerow}\textbf{0.627} & \cellcolor{bluerow}\underline{0.369} \\
    \midrule[\heavyrulewidth]
      \multirow[c]{6}{*}{ERNIE} & \multirow[c]{3}{*}{25\%} & REAP & \textbf{0.835} & \textbf{0.231} & 0.346 & 0.806 & \textbf{0.752} & 0.666 & 0.606 \\
    \cdashline{3-10}
       & & \cellcolor{bluerow}MAN & \cellcolor{bluerow}\underline{0.830} & \cellcolor{bluerow}\underline{0.214} & \cellcolor{bluerow}\underline{0.354} & \cellcolor{bluerow}\textbf{0.833} & \cellcolor{bluerow}\underline{0.740} & \cellcolor{bluerow}\underline{0.675} & \cellcolor{bluerow}\underline{0.608} \\
       & & \cellcolor{bluerow}MSAN & \cellcolor{bluerow}\textbf{0.835} & \cellcolor{bluerow}0.209 & \cellcolor{bluerow}\textbf{0.363} & \cellcolor{bluerow}\underline{0.831} & \cellcolor{bluerow}0.730 & \cellcolor{bluerow}\textbf{0.684} & \cellcolor{bluerow}\textbf{0.609} \\
    \cmidrule(lr){2-10}
       & \multirow[c]{3}{*}{50\%} & REAP & \textbf{0.737} & \underline{0.198} & \textbf{0.203} & 0.698 & \underline{0.582} & 0.578 & \underline{0.499} \\
    \cdashline{3-10}
       & & \cellcolor{bluerow}MAN & \cellcolor{bluerow}\underline{0.716} & \cellcolor{bluerow}\textbf{0.214} & \cellcolor{bluerow}0.167 & \cellcolor{bluerow}\textbf{0.751} & \cellcolor{bluerow}\underline{0.582} & \cellcolor{bluerow}\textbf{0.597} & \cellcolor{bluerow}\textbf{0.505} \\
       & & \cellcolor{bluerow}MSAN & \cellcolor{bluerow}0.692 & \cellcolor{bluerow}0.176 & \cellcolor{bluerow}\underline{0.199} & \cellcolor{bluerow}\underline{0.748} & \cellcolor{bluerow}\textbf{0.584} & \cellcolor{bluerow}\underline{0.592} & \cellcolor{bluerow}0.498 \\
    \midrule[\heavyrulewidth]
      \multirow[c]{6}{*}{Qwen3} & \multirow[c]{3}{*}{25\%} & REAP & 0.852 & 0.335 & \underline{0.572} & 0.903 & \underline{0.784} & 0.694 & 0.690 \\
    \cdashline{3-10}
       & & \cellcolor{bluerow}MAN & \cellcolor{bluerow}\textbf{0.871} & \cellcolor{bluerow}\textbf{0.396} & \cellcolor{bluerow}0.559 & \cellcolor{bluerow}\underline{0.910} & \cellcolor{bluerow}\textbf{0.800} & \cellcolor{bluerow}\underline{0.707} & \cellcolor{bluerow}\textbf{0.707} \\
       & & \cellcolor{bluerow}MSAN & \cellcolor{bluerow}\underline{0.860} & \cellcolor{bluerow}\underline{0.363} & \cellcolor{bluerow}\textbf{0.586} & \cellcolor{bluerow}\textbf{0.916} & \cellcolor{bluerow}0.778 & \cellcolor{bluerow}\textbf{0.709} & \cellcolor{bluerow}\underline{0.702} \\
    \cmidrule(lr){2-10}
       & \multirow[c]{3}{*}{50\%} & REAP & \underline{0.835} & \textbf{0.357} & 0.263 & 0.866 & \textbf{0.800} & 0.563 & 0.614 \\
    \cdashline{3-10}
       & & \cellcolor{bluerow}MAN & \cellcolor{bluerow}\textbf{0.850} & \cellcolor{bluerow}\underline{0.346} & \cellcolor{bluerow}\underline{0.275} & \cellcolor{bluerow}\textbf{0.884} & \cellcolor{bluerow}\underline{0.794} & \cellcolor{bluerow}\underline{0.612} & \cellcolor{bluerow}\textbf{0.627} \\
       & & \cellcolor{bluerow}MSAN & \cellcolor{bluerow}0.832 & \cellcolor{bluerow}0.324 & \cellcolor{bluerow}\textbf{0.280} & \cellcolor{bluerow}\underline{0.882} & \cellcolor{bluerow}0.792 & \cellcolor{bluerow}\textbf{0.614} & \cellcolor{bluerow}\underline{0.621} \\
    \bottomrule
  \end{tabular}%
  }%
  }
  \captionof{table}{\textbf{Robustness to the calibration distribution.} Results under Evol-CodeAlpaca-v1 calibration across all 16 benchmarks. Eval+ and MC Avg aggregate four EvalPlus and eight multiple-choice benchmarks; Avg averages the six displayed columns. Bold and underline denote the best and second-best result within each model--ratio block.}
  \label{tab:appendix_task_agnostic_evol_codealpaca}
}
\end{minipage}
}]
\clearpage

\twocolumn[{%
\begin{minipage}{\textwidth}
\captionsetup{hypcap=false}
\section{Detailed Task-Agnostic Results}
\label{sec:appendix_task_agnostic_results}
{\providecommand{\resulttablesep}{{\color{black!55}\vrule width 0.3pt}}
\global\long\def\taskagnosticdetailedrows{%
    \midrule
      \multirow[c]{8}{*}{OLMoE} & 0\% & Full & - & 0.354 & 0.323 & 0.373 & 0.312 & 0.033 & 0.444 & 0.682 & 0.222 & 0.490 & 0.758 & 0.766 & 0.808 & 0.534 & 0.470 & 0.711 & 0.684 \\
    \cmidrule(lr){2-20}
       & \multirow[c]{7}{*}{25\%} & Frequency & $(0,0,0)$ & 0.000 & 0.000 & 0.000 & 0.000 & 0.000 & 0.127 & 0.033 & 0.024 & 0.382 & 0.582 & 0.691 & 0.727 & 0.325 & 0.416 & \underline{0.690} & \textbf{0.665} \\
       &  & SEER & $(0,1,0)$ & 0.000 & 0.000 & 0.000 & 0.000 & 0.000 & 0.141 & 0.037 & 0.012 & 0.388 & 0.590 & 0.694 & 0.733 & 0.320 & 0.414 & \textbf{0.711} & \underline{0.661} \\
       &  & EAN & $(0,0,1)$ & 0.000 & 0.000 & 0.000 & 0.000 & 0.000 & 0.184 & 0.133 & 0.012 & 0.454 & 0.656 & \textbf{0.736} & 0.752 & 0.344 & \underline{0.422} & 0.632 & \underline{0.661} \\
       &  & REAP & $(1,1,1)$ & 0.000 & 0.000 & 0.000 & 0.000 & 0.000 & \textbf{0.263} & 0.139 & 0.036 & \textbf{0.487} & \textbf{0.720} & 0.695 & \textbf{0.760} & \textbf{0.397} & \textbf{0.436} & 0.661 & 0.654 \\
       &  & MoNE & - & 0.000 & 0.000 & 0.000 & 0.000 & 0.000 & 0.181 & 0.117 & 0.006 & 0.446 & 0.661 & \underline{0.727} & 0.751 & 0.351 & \textbf{0.436} & 0.639 & 0.653 \\
    \cdashline{3-20}
       &  & \cellcolor{bluerow}MAN & \cellcolor{bluerow}$(1,0,1)$ & \cellcolor{bluerow}\textbf{0.012} & \cellcolor{bluerow}\textbf{0.012} & \cellcolor{bluerow}\underline{0.005} & \cellcolor{bluerow}\underline{0.005} & \cellcolor{bluerow}0.000 & \cellcolor{bluerow}\underline{0.260} & \cellcolor{bluerow}\textbf{0.208} & \cellcolor{bluerow}\underline{0.046} & \cellcolor{bluerow}\underline{0.469} & \cellcolor{bluerow}\underline{0.683} & \cellcolor{bluerow}0.719 & \cellcolor{bluerow}\textbf{0.760} & \cellcolor{bluerow}\underline{0.395} & \cellcolor{bluerow}\textbf{0.436} & \cellcolor{bluerow}0.639 & \cellcolor{bluerow}0.653 \\
       &  & \cellcolor{bluerow}MSAN & \cellcolor{bluerow}$(1,0,2)$ & \cellcolor{bluerow}\underline{0.006} & \cellcolor{bluerow}\underline{0.006} & \cellcolor{bluerow}\textbf{0.011} & \cellcolor{bluerow}\textbf{0.011} & \cellcolor{bluerow}0.000 & \cellcolor{bluerow}0.242 & \cellcolor{bluerow}\underline{0.194} & \cellcolor{bluerow}\textbf{0.056} & \cellcolor{bluerow}0.457 & \cellcolor{bluerow}0.669 & \cellcolor{bluerow}0.692 & \cellcolor{bluerow}\underline{0.755} & \cellcolor{bluerow}\underline{0.395} & \cellcolor{bluerow}\textbf{0.436} & \cellcolor{bluerow}0.664 & \cellcolor{bluerow}0.646 \\
    \midrule[\heavyrulewidth]
      \multirow[c]{8}{*}{DeepSeek} & 0\% & Full & - & 0.591 & 0.524 & 0.585 & 0.497 & 0.104 & 0.418 & 0.610 & 0.298 & 0.541 & 0.785 & 0.829 & 0.808 & 0.567 & 0.456 & 0.726 & 0.712 \\
    \cmidrule(lr){2-20}
       & \multirow[c]{7}{*}{25\%} & Frequency & $(0,0,0)$ & 0.000 & 0.000 & 0.000 & 0.000 & 0.000 & \underline{0.291} & 0.023 & 0.012 & 0.428 & 0.667 & 0.729 & 0.751 & 0.417 & 0.424 & \textbf{0.726} & 0.671 \\
       &  & SEER & $(0,1,0)$ & 0.000 & 0.000 & 0.000 & 0.000 & 0.000 & 0.155 & 0.034 & 0.016 & 0.451 & 0.680 & 0.722 & 0.754 & 0.394 & 0.422 & \textbf{0.726} & 0.669 \\
       &  & EAN & $(0,0,1)$ & 0.000 & 0.000 & 0.000 & 0.000 & 0.000 & \textbf{0.295} & \underline{0.312} & 0.028 & 0.474 & 0.694 & 0.747 & \underline{0.779} & \underline{0.456} & 0.430 & 0.690 & \underline{0.703} \\
       &  & REAP & $(1,1,1)$ & \underline{0.006} & \underline{0.006} & \underline{0.008} & \underline{0.008} & 0.000 & 0.174 & 0.281 & 0.028 & 0.483 & 0.730 & 0.688 & \textbf{0.780} & \textbf{0.466} & 0.430 & 0.690 & 0.696 \\
       &  & MoNE & - & 0.000 & 0.000 & 0.000 & 0.000 & 0.000 & 0.200 & 0.227 & 0.024 & 0.479 & 0.688 & 0.750 & \underline{0.779} & 0.454 & \textbf{0.462} & \textbf{0.726} & 0.691 \\
    \cdashline{3-20}
       &  & \cellcolor{bluerow}MAN & \cellcolor{bluerow}$(1,0,1)$ & \cellcolor{bluerow}0.000 & \cellcolor{bluerow}0.000 & \cellcolor{bluerow}0.003 & \cellcolor{bluerow}0.003 & \cellcolor{bluerow}0.000 & \cellcolor{bluerow}0.154 & \cellcolor{bluerow}0.287 & \cellcolor{bluerow}\underline{0.032} & \cellcolor{bluerow}\underline{0.492} & \cellcolor{bluerow}\textbf{0.752} & \cellcolor{bluerow}\underline{0.764} & \cellcolor{bluerow}0.778 & \cellcolor{bluerow}0.454 & \cellcolor{bluerow}\underline{0.446} & \cellcolor{bluerow}\underline{0.693} & \cellcolor{bluerow}\textbf{0.704} \\
       &  & \cellcolor{bluerow}MSAN & \cellcolor{bluerow}$(1,0,2)$ & \cellcolor{bluerow}\textbf{0.012} & \cellcolor{bluerow}\textbf{0.012} & \cellcolor{bluerow}\textbf{0.040} & \cellcolor{bluerow}\textbf{0.037} & \cellcolor{bluerow}0.000 & \cellcolor{bluerow}0.238 & \cellcolor{bluerow}\textbf{0.428} & \cellcolor{bluerow}\textbf{0.102} & \cellcolor{bluerow}\textbf{0.503} & \cellcolor{bluerow}\underline{0.734} & \cellcolor{bluerow}\textbf{0.765} & \cellcolor{bluerow}0.778 & \cellcolor{bluerow}0.452 & \cellcolor{bluerow}\underline{0.446} & \cellcolor{bluerow}0.661 & \cellcolor{bluerow}\underline{0.703} \\
    \midrule[\heavyrulewidth]
      \multirow[c]{15}{*}{ERNIE} & 0\% & Full & - & 0.909 & 0.878 & 0.915 & 0.765 & 0.247 & 0.479 & 0.829 & 0.780 & 0.564 & 0.782 & 0.872 & 0.814 & 0.739 & 0.462 & 0.816 & 0.717 \\
    \cmidrule(lr){2-20}
       & \multirow[c]{7}{*}{25\%} & Frequency & $(0,0,0)$ & 0.201 & 0.165 & 0.360 & 0.288 & 0.055 & 0.352 & 0.647 & 0.316 & 0.518 & 0.727 & 0.849 & 0.719 & 0.571 & 0.390 & 0.791 & 0.679 \\
       &  & SEER & $(0,1,0)$ & 0.232 & 0.195 & 0.317 & 0.278 & 0.060 & 0.381 & 0.748 & 0.368 & 0.511 & 0.750 & 0.845 & 0.736 & 0.599 & 0.400 & 0.747 & 0.676 \\
       &  & EAN & $(0,0,1)$ & \underline{0.299} & \underline{0.262} & 0.347 & 0.294 & 0.055 & 0.408 & 0.673 & 0.370 & 0.535 & 0.750 & 0.840 & \textbf{0.790} & 0.597 & 0.436 & 0.791 & \textbf{0.716} \\
       &  & REAP & $(1,1,1)$ & 0.244 & 0.232 & 0.341 & 0.291 & 0.060 & \underline{0.414} & 0.760 & 0.528 & \textbf{0.577} & 0.791 & 0.851 & 0.775 & 0.645 & 0.444 & \underline{0.816} & \underline{0.704} \\
       &  & MoNE & - & 0.201 & 0.177 & 0.257 & 0.233 & 0.055 & \textbf{0.419} & 0.774 & 0.384 & 0.542 & 0.748 & 0.864 & \underline{0.777} & 0.540 & 0.442 & 0.812 & 0.701 \\
    \cdashline{3-20}
       &  & \cellcolor{bluerow}MAN & \cellcolor{bluerow}$(1,0,1)$ & \cellcolor{bluerow}0.293 & \cellcolor{bluerow}\underline{0.262} & \cellcolor{bluerow}\textbf{0.437} & \cellcolor{bluerow}\textbf{0.381} & \cellcolor{bluerow}\textbf{0.077} & \cellcolor{bluerow}0.402 & \cellcolor{bluerow}\underline{0.801} & \cellcolor{bluerow}\textbf{0.580} & \cellcolor{bluerow}\underline{0.573} & \cellcolor{bluerow}\textbf{0.798} & \cellcolor{bluerow}\underline{0.868} & \cellcolor{bluerow}0.776 & \cellcolor{bluerow}\underline{0.661} & \cellcolor{bluerow}\textbf{0.478} & \cellcolor{bluerow}0.794 & \cellcolor{bluerow}0.688 \\
       &  & \cellcolor{bluerow}MSAN & \cellcolor{bluerow}$(1,0,2)$ & \cellcolor{bluerow}\textbf{0.341} & \cellcolor{bluerow}\textbf{0.305} & \cellcolor{bluerow}\underline{0.410} & \cellcolor{bluerow}\underline{0.360} & \cellcolor{bluerow}\underline{0.066} & \cellcolor{bluerow}0.404 & \cellcolor{bluerow}\textbf{0.813} & \cellcolor{bluerow}\underline{0.564} & \cellcolor{bluerow}0.562 & \cellcolor{bluerow}\underline{0.793} & \cellcolor{bluerow}\textbf{0.874} & \cellcolor{bluerow}0.773 & \cellcolor{bluerow}\textbf{0.674} & \cellcolor{bluerow}\underline{0.458} & \cellcolor{bluerow}\textbf{0.820} & \cellcolor{bluerow}0.678 \\
    \cmidrule(lr){2-20}
       & \multirow[c]{7}{*}{50\%} & Frequency & $(0,0,0)$ & 0.012 & 0.012 & 0.003 & 0.003 & 0.000 & 0.164 & 0.051 & 0.018 & 0.386 & 0.583 & 0.740 & 0.579 & 0.488 & 0.324 & 0.733 & 0.625 \\
       &  & SEER & $(0,1,0)$ & 0.000 & 0.000 & 0.005 & 0.005 & 0.000 & 0.174 & 0.106 & 0.020 & 0.386 & 0.599 & 0.711 & 0.573 & 0.460 & 0.324 & 0.675 & 0.613 \\
       &  & EAN & $(0,0,1)$ & 0.006 & 0.006 & 0.013 & 0.013 & 0.005 & 0.256 & 0.080 & 0.026 & 0.446 & 0.675 & 0.732 & \textbf{0.680} & 0.470 & \underline{0.394} & \textbf{0.762} & \textbf{0.698} \\
       &  & REAP & $(1,1,1)$ & 0.012 & 0.012 & 0.011 & 0.008 & 0.000 & 0.237 & 0.478 & 0.158 & 0.402 & 0.596 & 0.736 & 0.666 & 0.394 & \textbf{0.398} & 0.715 & 0.676 \\
       &  & MoNE & - & 0.006 & 0.006 & 0.003 & 0.003 & 0.000 & 0.257 & 0.174 & 0.048 & 0.381 & 0.599 & \textbf{0.802} & \underline{0.677} & 0.376 & 0.390 & \underline{0.737} & \underline{0.682} \\
    \cdashline{3-20}
       &  & \cellcolor{bluerow}MAN & \cellcolor{bluerow}$(1,0,1)$ & \cellcolor{bluerow}\textbf{0.043} & \cellcolor{bluerow}\textbf{0.043} & \cellcolor{bluerow}\underline{0.077} & \cellcolor{bluerow}\underline{0.066} & \cellcolor{bluerow}\textbf{0.022} & \cellcolor{bluerow}\underline{0.273} & \cellcolor{bluerow}\underline{0.672} & \cellcolor{bluerow}\underline{0.190} & \cellcolor{bluerow}\underline{0.490} & \cellcolor{bluerow}\underline{0.714} & \cellcolor{bluerow}\underline{0.797} & \cellcolor{bluerow}\underline{0.677} & \cellcolor{bluerow}\textbf{0.543} & \cellcolor{bluerow}0.386 & \cellcolor{bluerow}0.733 & \cellcolor{bluerow}0.664 \\
       &  & \cellcolor{bluerow}MSAN & \cellcolor{bluerow}$(1,0,2)$ & \cellcolor{bluerow}\underline{0.037} & \cellcolor{bluerow}\underline{0.037} & \cellcolor{bluerow}\textbf{0.093} & \cellcolor{bluerow}\textbf{0.085} & \cellcolor{bluerow}\underline{0.011} & \cellcolor{bluerow}\textbf{0.285} & \cellcolor{bluerow}\textbf{0.681} & \cellcolor{bluerow}\textbf{0.200} & \cellcolor{bluerow}\textbf{0.526} & \cellcolor{bluerow}\textbf{0.737} & \cellcolor{bluerow}0.691 & \cellcolor{bluerow}\textbf{0.680} & \cellcolor{bluerow}\underline{0.500} & \cellcolor{bluerow}0.382 & \cellcolor{bluerow}0.675 & \cellcolor{bluerow}0.668 \\
    \midrule[\heavyrulewidth]
      \multirow[c]{15}{*}{Qwen3} & 0\% & Full & - & 0.939 & 0.902 & 0.892 & 0.751 & 0.368 & 0.644 & 0.923 & 0.802 & 0.625 & 0.838 & 0.887 & 0.797 & 0.802 & 0.446 & 0.769 & 0.736 \\
    \cmidrule(lr){2-20}
       & \multirow[c]{7}{*}{25\%} & Frequency & $(0,0,0)$ & 0.000 & 0.000 & 0.000 & 0.000 & 0.000 & \textbf{0.632} & 0.904 & 0.196 & 0.625 & 0.839 & \underline{0.886} & \textbf{0.795} & 0.761 & 0.442 & 0.773 & \underline{0.732} \\
       &  & SEER & $(0,1,0)$ & 0.006 & 0.006 & 0.000 & 0.000 & 0.000 & 0.612 & 0.913 & 0.202 & 0.630 & 0.845 & \textbf{0.888} & \textbf{0.795} & 0.760 & \underline{0.446} & 0.776 & \underline{0.732} \\
       &  & EAN & $(0,0,1)$ & 0.000 & 0.000 & 0.003 & 0.003 & 0.000 & 0.623 & 0.910 & 0.194 & \textbf{0.642} & \textbf{0.854} & 0.885 & \textbf{0.795} & \textbf{0.768} & 0.442 & 0.765 & 0.731 \\
       &  & REAP & $(1,1,1)$ & 0.585 & 0.549 & 0.683 & 0.579 & 0.137 & 0.600 & 0.879 & 0.778 & 0.620 & 0.825 & 0.885 & \underline{0.776} & 0.754 & 0.426 & 0.751 & 0.729 \\
       &  & MoNE & - & 0.000 & 0.000 & 0.000 & 0.000 & 0.000 & \underline{0.627} & 0.911 & 0.206 & 0.632 & \underline{0.851} & \underline{0.886} & \textbf{0.795} & \underline{0.766} & 0.438 & 0.765 & \textbf{0.734} \\
    \cdashline{3-20}
       &  & \cellcolor{bluerow}MAN & \cellcolor{bluerow}$(1,0,1)$ & \cellcolor{bluerow}\textbf{0.945} & \cellcolor{bluerow}\textbf{0.896} & \cellcolor{bluerow}\underline{0.889} & \cellcolor{bluerow}\underline{0.743} & \cellcolor{bluerow}\textbf{0.346} & \cellcolor{bluerow}0.570 & \cellcolor{bluerow}\textbf{0.937} & \cellcolor{bluerow}\underline{0.792} & \cellcolor{bluerow}\underline{0.641} & \cellcolor{bluerow}\underline{0.851} & \cellcolor{bluerow}0.877 & \cellcolor{bluerow}0.766 & \cellcolor{bluerow}0.736 & \cellcolor{bluerow}\textbf{0.448} & \cellcolor{bluerow}\underline{0.780} & \cellcolor{bluerow}0.717 \\
       &  & \cellcolor{bluerow}MSAN & \cellcolor{bluerow}$(1,0,2)$ & \cellcolor{bluerow}\underline{0.890} & \cellcolor{bluerow}\underline{0.848} & \cellcolor{bluerow}\textbf{0.894} & \cellcolor{bluerow}\textbf{0.757} & \cellcolor{bluerow}\underline{0.335} & \cellcolor{bluerow}0.559 & \cellcolor{bluerow}\underline{0.935} & \cellcolor{bluerow}\textbf{0.796} & \cellcolor{bluerow}0.636 & \cellcolor{bluerow}0.847 & \cellcolor{bluerow}0.872 & \cellcolor{bluerow}0.769 & \cellcolor{bluerow}0.743 & \cellcolor{bluerow}0.440 & \cellcolor{bluerow}\textbf{0.794} & \cellcolor{bluerow}0.718 \\
    \cmidrule(lr){2-20}
       & \multirow[c]{7}{*}{50\%} & Frequency & $(0,0,0)$ & 0.000 & 0.000 & 0.000 & 0.000 & 0.000 & 0.015 & 0.000 & 0.000 & 0.287 & 0.391 & 0.655 & 0.436 & 0.276 & 0.306 & 0.585 & 0.564 \\
       &  & SEER & $(0,1,0)$ & 0.000 & 0.000 & 0.000 & 0.000 & 0.000 & 0.018 & 0.000 & 0.000 & 0.288 & 0.396 & 0.651 & 0.435 & 0.276 & 0.302 & 0.567 & 0.571 \\
       &  & EAN & $(0,0,1)$ & 0.000 & 0.000 & 0.000 & 0.000 & 0.000 & \textbf{0.530} & 0.656 & 0.036 & \textbf{0.562} & \underline{0.766} & \textbf{0.885} & \textbf{0.785} & \underline{0.634} & \underline{0.438} & 0.773 & \textbf{0.734} \\
       &  & REAP & $(1,1,1)$ & \underline{0.006} & \underline{0.006} & 0.000 & 0.000 & 0.000 & 0.333 & 0.849 & 0.690 & 0.506 & 0.710 & \underline{0.865} & \underline{0.700} & 0.617 & 0.382 & \textbf{0.798} & \underline{0.701} \\
       &  & MoNE & - & 0.000 & 0.000 & 0.000 & 0.000 & 0.000 & \underline{0.454} & 0.632 & 0.016 & \underline{0.555} & \textbf{0.776} & \textbf{0.885} & \textbf{0.785} & \textbf{0.639} & \textbf{0.450} & \underline{0.776} & \textbf{0.734} \\
    \cdashline{3-20}
       &  & \cellcolor{bluerow}MAN & \cellcolor{bluerow}$(1,0,1)$ & \cellcolor{bluerow}0.000 & \cellcolor{bluerow}0.000 & \cellcolor{bluerow}\underline{0.021} & \cellcolor{bluerow}\underline{0.021} & \cellcolor{bluerow}\underline{0.027} & \cellcolor{bluerow}0.281 & \cellcolor{bluerow}\underline{0.898} & \cellcolor{bluerow}\textbf{0.792} & \cellcolor{bluerow}0.540 & \cellcolor{bluerow}0.747 & \cellcolor{bluerow}0.842 & \cellcolor{bluerow}0.616 & \cellcolor{bluerow}0.528 & \cellcolor{bluerow}0.368 & \cellcolor{bluerow}0.700 & \cellcolor{bluerow}0.665 \\
       &  & \cellcolor{bluerow}MSAN & \cellcolor{bluerow}$(1,0,2)$ & \cellcolor{bluerow}\textbf{0.256} & \cellcolor{bluerow}\textbf{0.250} & \cellcolor{bluerow}\textbf{0.370} & \cellcolor{bluerow}\textbf{0.328} & \cellcolor{bluerow}\textbf{0.060} & \cellcolor{bluerow}0.217 & \cellcolor{bluerow}\textbf{0.910} & \cellcolor{bluerow}\underline{0.784} & \cellcolor{bluerow}0.547 & \cellcolor{bluerow}0.755 & \cellcolor{bluerow}0.837 & \cellcolor{bluerow}0.611 & \cellcolor{bluerow}0.527 & \cellcolor{bluerow}0.376 & \cellcolor{bluerow}0.718 & \cellcolor{bluerow}0.672 \\
    \bottomrule
}

  \centering
  \scriptsize
  \setlength{\tabcolsep}{2pt}
  \resizebox{\textwidth}{!}{%
  \begin{tabular}{c c l c!{\resulttablesep}c c c c c!{\resulttablesep}c!{\resulttablesep}c c *{8}{@{}H@{}}}
    \toprule
    \multicolumn{4}{c!{\resulttablesep}}{} & \multicolumn{5}{c!{\resulttablesep}}{\textbf{Coding}} & \multicolumn{1}{c!{\resulttablesep}}{\textbf{Writing}} & \multicolumn{2}{c}{\textbf{Math}} & \multicolumn{8}{@{}c@{}}{} \\
    \textbf{Model} & \textbf{Pruning Ratio} & \textbf{Criterion} & \textbf{$(b,\alpha,\beta)$} & \textbf{HEVAL} & \textbf{HEVAL+} & \textbf{MBPP} & \textbf{MBPP+} & \textbf{LiveCode} & \textbf{WildBench} & \textbf{GSM8K} & \textbf{MATH-500} & \multicolumn{8}{@{}c@{}}{} \\
    \taskagnosticdetailedrows
  \end{tabular}%
  }
  \captionof{table}{\textbf{Detailed task-agnostic generative and reasoning results under C4 calibration.} OLMoE and DeepSeek are reported at a 25\% pruning ratio; ERNIE and Qwen3 are reported at 25\% and 50\%. HumanEval and HumanEval+ are abbreviated as HEVAL and HEVAL+. HEVAL, HEVAL+, MBPP, and MBPP+ are the Eval+ sub-benchmarks.}
  \label{tab:pvar_task_agnostic_c4_appendix}
}
\end{minipage}
}]
\clearpage

\twocolumn[{%
\begin{minipage}{\textwidth}
\captionsetup{hypcap=false}
{  \centering
  \scriptsize
  \setlength{\tabcolsep}{3pt}
  \resizebox{0.96\textwidth}{!}{%
  \begin{tabular}{c c l c *{8}{@{}H@{}} !{\resulttablesep}c c c c c c c c}
    \toprule
    \multicolumn{4}{c}{} & \multicolumn{8}{@{}c@{}}{} & \multicolumn{8}{c}{\textbf{Multiple Choice}} \\
    \textbf{Model} & \textbf{Pruning Ratio} & \textbf{Criterion} & \textbf{$(b,\alpha,\beta)$} & \multicolumn{8}{@{}c@{}}{} & \textbf{ARC-C} & \textbf{ARC-E} & \textbf{BoolQ} & \textbf{HellaSwag} & \textbf{MMLU} & \textbf{OBQA} & \textbf{RTE} & \textbf{WinoG} \\
    \taskagnosticdetailedrows
  \end{tabular}%
  }
  \captionof{table}{\textbf{Detailed task-agnostic multiple-choice results under C4 calibration.} Pruning ratios match \Cref{tab:pvar_task_agnostic_c4_appendix}. OpenBookQA and WinoGrande are abbreviated as OBQA and WinoG., respectively.}
  \label{tab:pvar_task_agnostic_c4_appendix_mc}
}
\end{minipage}
}]
\clearpage

\twocolumn[{%
\begin{minipage}{\textwidth}
\captionsetup{hypcap=false}
\section{Score Definitions in the Systematic Formulation}
\label{sec:appendix_metric_scores}
{
\centering
\small
\setlength{\tabcolsep}{6pt}
\renewcommand{\arraystretch}{1.1}
\resizebox{0.68\textwidth}{!}{%
\begin{tabular}{llc}
\toprule
Name & Formula & Hyperparameters $(b,\alpha,\beta)$ \\
\midrule
Frequency & $\sum_{t=1}^{M}\mathbf{1}[j\in\mathcal{E}_t]$ & $(0,0,0)$ \\
SEER & $\sum_{t=1}^{M}\mathbf{1}[j\in\mathcal{E}_t]g_{j,t}$ & $(0,1,0)$ \\
EAN & $\sum_{t=1}^{M}\mathbf{1}[j\in\mathcal{E}_t]\|\mathbf{f}_{j,t}\|_2$ & $(0,0,1)$ \\
REAP & $\frac{1}{N_j}\sum_{t=1}^{M}\mathbf{1}[j\in\mathcal{E}_t]g_{j,t}\|\mathbf{f}_{j,t}\|_2$ & $(1,1,1)$ \\
MoNE & $\left(\frac{1}{N_j}\sum_{t=1}^{M}\mathbf{1}[j\in\mathcal{E}_t]g_{j,t}\right)\mathrm{Var}_j(\mathbf{f})$ & - \\
MAN & $\frac{1}{N_j}\sum_{t=1}^{M}\mathbf{1}[j\in\mathcal{E}_t]\|\mathbf{f}_{j,t}\|_2$ & $(1,0,1)$ \\
MSAN & $\frac{1}{N_j}\sum_{t=1}^{M}\mathbf{1}[j\in\mathcal{E}_t]\|\mathbf{f}_{j,t}\|_2^{2}$ & $(1,0,2)$ \\
RCN & $\sum_{t=1}^{M}\mathbf{1}[j\in\mathcal{E}_t]g_{j,t}\|\mathbf{f}_{j,t}\|_2$ & $(0,1,1)$ \\
RCE & $\sum_{t=1}^{M}\mathbf{1}[j\in\mathcal{E}_t]g_{j,t}^{2}\|\mathbf{f}_{j,t}\|_2^{2}$ & $(0,2,2)$ \\
\bottomrule
\end{tabular}
}
\captionof{table}{Score definitions for the scoring criteria in the systematic formulation. Here \(N_j=\sum_{t=1}^{M}\mathbf{1}[j\in\mathcal{E}_t]\). For MoNE, \(\mathrm{Var}_j(\mathbf{f})=\left\|\sqrt{\frac{1}{N_j-1}\sum_{t=1}^{M}\mathbf{1}[j\in\mathcal{E}_t](\mathbf{f}_{j,t}-\bar{\mathbf{f}}_j)^2}\right\|_2\), where \(\bar{\mathbf{f}}_j=\frac{1}{N_j}\sum_{t=1}^{M}\mathbf{1}[j\in\mathcal{E}_t]\mathbf{f}_{j,t}\), and the square and square root are elementwise.}
\label{tab:appendix_score_family}
}
\end{minipage}
}]

\end{document}